\documentclass[twoside,11pt]{article}

\usepackage{amsfonts}
\usepackage{mathrsfs}
\usepackage{dsfont}
\usepackage[utf8]{inputenc} 
\usepackage{float}
\usepackage{enumitem}
\usepackage{amsmath,amssymb}
\usepackage{latexsym}
\usepackage{longtable}
\usepackage{booktabs}
\usepackage{epstopdf}
\usepackage{tabularx}
\usepackage{tikz}
\usetikzlibrary{shapes.geometric, arrows}
\tikzstyle{startstop} = [rectangle, rounded corners, minimum width=3cm, minimum height=1cm, text centered, draw=black, fill=red!30]
\tikzstyle{process}   = [rectangle, minimum width=3cm, minimum height=1cm, text centered, draw=black, fill=blue!30]
\tikzstyle{io}        = [trapezium, trapezium left angle=70, trapezium right angle=110, minimum width=3cm, minimum height=1cm, text centered, draw=black, fill=green!30]
\tikzstyle{arrow}     = [thick,->,>=stealth]
\usepackage{natbib}
\usepackage{geometry}  
\usepackage[active]{srcltx}
\usepackage{graphicx}
\usepackage{booktabs}
\usepackage{multirow}
\usepackage{siunitx}
\usepackage{xurl} 
\usepackage{amsmath}
\usepackage{setspace}
\usepackage{algorithm}
\usepackage{algorithmic}
\usepackage{subcaption} 

\usepackage[
    bookmarks=true,
    bookmarksnumbered=true,
    colorlinks=true,
    pdfstartview=FitV,
    linkcolor=blue,
    citecolor=blue,
    urlcolor=blue
]{hyperref}

\usepackage{fancyhdr}
\usepackage{lipsum}
\usepackage{listings}
\usepackage{xcolor}

\definecolor{codegreen}{rgb}{0,0.6,0}
\definecolor{codegray} {rgb}{0.5,0.5,0.5}
\definecolor{codepurple}{rgb}{0.58,0,0.82}
\definecolor{backcolour}{rgb}{0.95,0.95,0.92}
\lstdefinestyle{mystyle}{
    backgroundcolor=\color{backcolour},
    commentstyle=\color{codegreen},
    keywordstyle=\color{magenta},
    numberstyle=\tiny\color{codegray},
    stringstyle=\color{codepurple},
    basicstyle=\footnotesize\ttfamily,
    breakatwhitespace=false,
    breaklines=true,
    captionpos=b,
    keepspaces=true,
    numbers=left,
    numbersep=5pt,
    showspaces=false,
    showstringspaces=false,
    showtabs=false,
    tabsize=2
}
\providecommand{\U}[1]{\protect\rule{.1in}{.1in}}
\newtheorem {theorem}{Theorem}[section]

\usepackage{titlesec}
\titleformat{\section}{\normalfont\Large\bfseries}{\thesection.}{1em}{}

\begin{document}

\title{\Large\textbf{Temporal Conformal Prediction (TCP):
Rolling Calibration for Financial Risk Forecasting}}

\author{
Agnideep Aich\textsuperscript{1,2},
Ashit Baran Aich\textsuperscript{3}\thanks{
Corresponding author: Ashit Baran Aich,
\href{mailto:aichnsou@gmail.com}{\texttt{aichnsou@gmail.com}}.
AI assisted with manuscript drafting and editing. The authors are responsible for the final manuscript and any remaining errors.
}
and Dipak C. Jain\textsuperscript{4}
\\[1ex]
\textsuperscript{1}Department of Mathematics,
University of Louisiana at Lafayette,\\
Lafayette, Louisiana, USA
\\[0.5ex]
\textsuperscript{2}Current affiliation:
Department of Emergency Medicine,\\
Stanford University, Palo Alto, California, USA
\\[0.5ex]
\textsuperscript{3}Formerly of the Department of Statistics,
Presidency College,\\
Kolkata, India
\\[0.5ex]
\textsuperscript{4}Department of Marketing,
China Europe International Business School (CEIBS),\\
Shanghai, PR China
}

\date{}
\maketitle
\vspace{-20pt}

\begin{abstract}
We study two rolling conformal procedures that construct prediction intervals for financial returns. Temporal Conformal Prediction (TCP) uses a zero-truncated version of the conformalized quantile regression (CQR) score. TCP-RM adds an offset updated from observed coverage errors with a decaying step size. We compare ten forecasting methods and two diagnostic controls across thirteen financial series on 1,448 common forecast dates. The nominal interval coverage is 95\%. TCP attains coverage of 95.23\% for the S\&P 500, 95.37\% for Bitcoin and 94.27\% for Gold. Across all thirteen series, CQR has the smallest mean absolute deviation from the coverage target, followed by TCP and TCP-RM. TCP and CQR produce different intervals on 3.655\% of forecasts. TCP-RM changes coverage and width little at the main settings. Coverage and interval score give different rankings. In 52 paired interval-score comparisons between TCP and four volatility-based benchmarks, 42 favour the volatility model and ten show no significant difference. These are individual Diebold--Mariano tests at the 5\% level without a multiple-testing adjustment. Crisis-window results vary across markets, and good full-sample interval coverage does not ensure lower-tail calibration. The standard split-conformal guarantee requires exchangeability and does not automatically apply to the dependent returns evaluated here. The results support evaluating rolling conformal forecasts through coverage, interval scores and separate tail diagnostics, rather than closeness to the coverage target alone.
\end{abstract}



\section{Introduction}
\label{sec:introduction}

Financial returns exhibit changing volatility and heavy tails \citep{Cont2001}. These features make prediction intervals difficult to calibrate. An interval may attain its coverage target over a long sample yet miss many observations during a short period of market stress. Width also matters: an interval that covers most returns can still give an imprecise forecast. We therefore assess coverage, interval scores and the timing of lower-tail exceptions separately.

Conformal prediction calibrates prediction sets using past forecast errors. Its standard finite-sample guarantee requires exchangeability \citep{Vovk2005}. Conformalized quantile regression (CQR) applies this calibration to estimated conditional quantiles \citep{Romano2019}. Rolling training and calibration windows allow the forecasts to change over time. They do not, by themselves, establish exchangeability or guarantee coverage for dependent returns. Extensions to non-exchangeable data require their own assumptions and coverage analysis \citep{BarberCandesRamdasTibshirani2023,Oliveira2024}.

We study two rolling constructions for financial returns. Temporal Conformal Prediction (TCP) uses the CQR score truncated at zero. This prevents TCP from narrowing the fitted quantile interval. TCP-RM adds an offset updated from observed coverage errors, with a decaying step size and a projection. Its unprojected increment has the same form as existing quantile-tracking updates \citep{Angelopoulos2023}. Decaying step sizes have also been studied for online conformal prediction \citep{AngelopoulosBarberBates2024}. Our focus is the behaviour of these constructions in a common empirical comparison. The offset update and the use of a decaying step size are not new mechanisms.

The study covers thirteen series spanning equity indices, cryptocurrencies, commodities and exchange rates. We compare ten forecasting methods and two diagnostic controls on 1,448 common forecast dates, from 23 January 2019 to 20 May 2025. The main interval target is 95\% coverage. The comparisons include rolling CQR, a clipped fixed-step ACI implementation, the uncalibrated quantile forecaster, fitted GARCH-t, filtered historical simulation, a filtered extreme-value construction, RiskMetrics and historical simulation. The S\&P 500, Bitcoin and Gold provide the main examples. The appendix reports results across all thirteen series, crisis-window results for Bitcoin and Gold, and sensitivity checks.

Full-sample coverage is close to the target for CQR, TCP and TCP-RM. CQR has the smallest mean absolute coverage deviation across the thirteen series, followed by TCP and TCP-RM. TCP and CQR differ on only 688 of 18,824 forecasts. The online offset changes little at the main settings. Coverage and interval score give different rankings. In 52 paired interval-score comparisons against GARCH-t, filtered historical simulation, GARCH-EVT and RiskMetrics, 42 favour the volatility model and ten show no significant difference. None favours TCP. These are individual Diebold--Mariano tests at the 5\% level, without a multiple-testing adjustment.

Short-window results also differ across markets. During February--April 2020, TCP covers 80.00\% of S\&P 500 returns, compared with 94.55\% for GARCH-t. On Gold, TCP covers 87.27\%, compared with 70.91\% for GARCH-t. These observations do not support a single crisis-period ranking. Lower-tail tests provide a further distinction: good two-sided coverage does not ensure a calibrated lower endpoint. We therefore treat those endpoints as diagnostics and evaluate separately constructed one-sided conformal forecasts. Neither TCP nor TCP-RM supplies an expected-shortfall forecast in this study.

Section~\ref{sec:related} reviews the relevant literature. Section~\ref{sec:notation} defines the notation. Section~\ref{sec:math} describes the constructions and the assumptions behind the rank guarantee. Section~\ref{sec:methodology} gives the empirical design, and Section~\ref{sec:result} reports the results. Section~\ref{sec:conclusion} discusses limitations and future work. Appendices~\ref{app:math}--\ref{app:traces} contain the proofs, further results and forecast traces.

\section{Related Work}
\label{sec:related}

\paragraph{Financial return and tail-risk forecasts.}
GARCH models describe changes in conditional variance \citep{Bollerslev1986}. RiskMetrics uses an exponentially weighted variance estimate \citep{Morgan1996}. Filtered historical simulation combines volatility filtering with an empirical residual distribution \citep{BaroneAdesi1999}. Filtered extreme-value methods model the tail of the standardised residuals \citep{McNeil2000}. These approaches provide different ways to construct return quantiles. Their performance depends on the variance model, the residual distribution and the estimation scheme. We compare fitted and filtered versions on the same forecast dates. Our GARCH-EVT intervals use an EVT lower endpoint and an FHS upper endpoint, as specified in Section~\ref{sec:methodology}.

Interval coverage and tail-risk evaluation answer different questions. The unconditional-coverage test checks the exception rate \citep{Kupiec1995}. Conditional-coverage and dynamic-quantile tests also assess aspects of the exception sequence \citep{Christoffersen1998,EngleManganelli2004}. Joint scores for value-at-risk and expected shortfall require forecasts of both quantities \citep{Patton2019}. A conformal interval alone does not supply an expected-shortfall estimate. These distinctions guide our evaluation of the interval and lower-tail forecasts.

\paragraph{Quantile forecasting and conformal calibration.}
Quantile regression estimates conditional quantiles directly \citep{Koenker1978}. Our base forecasts use LightGBM quantile regressors \citep{Ke2017}. Neural forecasting methods offer other model choices, as reviewed by \citet{Benidis2022}. Pretrained probabilistic models such as Chronos learn from collections of time series \citep{Ansari2024}. We use one fixed tree-based configuration and one seed. The experiments therefore assess calibration around that forecaster; they do not establish a ranking against neural or pretrained models.

CQR combines fitted quantiles with conformal calibration \citep{Romano2019}. It is the closest construction to TCP. With the same fitted quantiles, calibration observations and order-statistic rule, the TCP threshold is the maximum of the CQR threshold and zero. The two procedures coincide whenever the CQR threshold is non-negative. Truncation prevents contraction for TCP, but the TCP-RM offset can still make its total expansion negative. The general conformal rank argument comes from the conformal prediction literature \citep{Vovk2005}; using a rolling window does not create a separate validity result.

\paragraph{Conformal prediction with temporal dependence.}
Several approaches address the failure of exchangeability in time series. \citet{BarberCandesRamdasTibshirani2023} develop weighted conformal procedures and bounds on coverage loss under non-exchangeability. \citet{Oliveira2024} give a split-conformal analysis based on concentration and decoupling conditions, with a coverage penalty for dependence. \citet{BarberPananjady2025} study split conformal prediction with predictors that use past observations. They bound coverage loss through a switch coefficient and analyse stationary mixing processes. These results show how dependence can enter a coverage bound. We do not verify their assumptions or estimate their bounds for the return series in this study.

Other methods explicitly model dependence in the calibration scores. Sequential predictive conformal inference estimates a future score quantile conditional on past information \citep{XuXie2023}. A recent survey by \citet{Stocker2025} compares approaches based on reweighting, updating residual distributions and online adaptation. Our rolling calibration uses an empirical score quantile within a recent window. It does not fit a separate model for the score distribution.

\paragraph{Online adaptation.}
Adaptive conformal inference (ACI) updates the nominal miscoverage parameter in response to observed errors \citep{Gibbs2021}. \citet{zaffran2022adaptive} study its learning-rate sensitivity and introduce AgACI, which aggregates ACI forecasts across learning rates. \citet{GibbsCandes2024} develop an online procedure that adapts the step size and controls regret over local time intervals. \citet{BhatnagarWangXiongBai2023} use strongly adaptive online learning to obtain regret guarantees across time intervals of different lengths. Our empirical ACI comparator uses clipping and one fixed step size. Its results do not represent a comparison with those aggregated or adaptively tuned procedures.

Direct threshold updates also have established precedents. The quantile-tracking component of conformal PID control updates a score threshold using coverage errors \citep{Angelopoulos2023}. \citet{AngelopoulosBarberBates2024} study decaying step sizes, relating retrospective coverage to quantile estimation under stable distributions. TCP-RM applies an error-based offset to a newly computed rolling threshold and uses a moving projection bound. Those implementation choices require a separate analysis. We establish finite-horizon offset bounds in Appendix~\ref{app:rm-properties}; we do not claim that the earlier online coverage guarantees automatically apply.

\paragraph{Conformal methods in financial risk studies.}
Conformal methods have already been applied to financial returns and market risk. The conformal PID study includes stock-return experiments \citep{Angelopoulos2023}. \citet{Fantazzini2024} evaluates four adaptive conformal algorithms for VaR estimation on four thousand crypto-assets, alongside volatility-based benchmarks. Our study examines a narrower set of conformal implementations across several asset classes. It compares full-sample coverage, interval scores, short-window behaviour and lower-tail diagnostics under a common forecast protocol. These comparisons show why closeness to a nominal coverage target should be assessed alongside interval width, missed-return penalties and the timing of exceptions.

\section{Notation}
\label{sec:notation}

\paragraph{Returns and forecasts.}
The index $t$ counts recorded observations, which need not fall on consecutive calendar days. We write $r_t$ for the return stored at observation $t$ and $X_t$ for the features available before that return is observed. The forecast interval is $I_t=[\ell_t,u_t]$, with target coverage $1-\alpha$. The main analysis uses $\alpha=0.05$. Returns, interval endpoints and calibration offsets use the stored percentage scale. We use $T$ for the number of evaluated forecasts; tables report sample counts as $n$.

\paragraph{Training and calibration.}
The total rolling window has size $w$. Its training and calibration index sets are $\mathcal{T}_t$ and $\mathcal{C}_t$, with sizes $w-m$ and $m$, respectively. For any quantile level $\tau\in(0,1)$, $\widehat q_{\tau,t}(x)$ denotes the fitted quantile function at forecast origin $t$. In general conformal statements, $(X_i,Y_i)$ denotes a feature--outcome pair. For the financial application, the outcome is $Y_i=r_i$.

We write $s_{i,t}$ for the signed CQR calibration score and
\[
s^+_{i,t}=\max\{s_{i,t},0\}
\]
for its zero-truncated TCP counterpart. Their calibration thresholds are $c_t^{\mathrm{CQR}}$ and $c_t$. A parenthesised subscript, as in $s_{(k),t}$, denotes an order statistic. The score formulas and threshold conventions are given in Section~\ref{sec:math}. The set $\mathcal{C}_t$ and the scalar threshold $c_t$ are distinct objects.

\paragraph{Online calibration.}
For the interval being evaluated, the miss indicator is
\[
e_t=\mathbf{1}\{r_t\notin I_t\}.
\]
The centred miscoverage error is $e_t-\alpha$. A superscript identifies the method when needed, as in $e_t^{\mathrm{ACI}}$.

TCP-RM adds an offset $b_t$ to the TCP threshold, giving total expansion $c_t+b_t$. The first forecast index is $t_0$, and $j(t)=t-t_0+1$ indexes the updates. The step size $\gamma_j$ is controlled by $\gamma_0$, $\lambda$ and $\beta$. For clipped ACI, $a_t$ denotes the updated miscoverage level, $h_t$ its interval expansion and $\eta$ its fixed step size. The update rules are defined in Sections~\ref{sec:adaptive} and \ref{sec:methodology}.

\paragraph{Lower-tail evaluation.}
In lower-tail evaluation, $\tau$ refers to the requested lower-tail level. In a backtest, $q_t$ is the lower endpoint being tested and $p$ is the reference exception probability. The exception indicator is
\[
H_t=\mathbf{1}\{r_t<q_t\}.
\]
Thus $H_t$ records a lower-tail exception, whereas $e_t$ records a miss on either side of an interval. We write $\widehat{\mathrm{ES}}_{\tau,t}$ for an expected-shortfall forecast. The joint-score formula uses $E_{\tau,t}$ as shorthand for this forecast.

\paragraph{Volatility and EVT quantities.}
The feature $v_{20,t}$ is the sample standard deviation of the preceding 20 recorded returns. In the fitted GARCH model, $\mu$ is the conditional mean and $\sigma_t^2$ is the conditional variance. Its variance parameters are $\omega$, $a_G$ and $b_G$, and $\nu$ is the Student-$t$ degrees of freedom. Hats identify fitted quantities. We use $v_t$ for the variance in the RiskMetrics and fixed-coefficient recursions.

In GARCH-EVT, $u_t^{\mathrm{EVT}}$ is the tail threshold and $v_{\tau,t}$ is the fitted loss quantile. Both are on the standardised loss scale. They are distinct from the upper return endpoint $u_t$, the volatility feature $v_{20,t}$ and the recursion variance $v_t$.

\paragraph{Operators.}
We use $\Pr(\cdot)$ for probability, $\mathbf{1}\{\cdot\}$ for an indicator and $\operatorname{sign}(\cdot)$ for the sign function. The operators $\lfloor\cdot\rfloor$ and $\lceil\cdot\rceil$ denote the floor and ceiling.

\section{Mathematical Background}
\label{sec:math}

We study rolling prediction intervals for financial returns. TCP uses a quantile forecaster followed by conformal calibration. Its calibration score is a zero-truncated version of the score used in conformalized quantile regression (CQR) \citep{Romano2019}. TCP-RM adds an offset that responds to past coverage errors. We distinguish these constructions from the assumptions needed for a coverage guarantee.

\subsection{Prediction target and notation}

Let $r_t$ denote the return at the $t$th recorded observation. Before observing $r_t$, we construct an interval $I_t=[\ell_t,u_t]$ using the available history. The feature vector $X_t$ contains only information available before that observation. The target coverage is $1-\alpha$, with $\alpha=0.05$ in the main analysis. This is a target for the forecasting procedure, not an assumed property of its output.

We use $\mathcal{T}_t$ for the training indices and $\mathcal{C}_t$ for the calibration indices. The scalar calibration threshold is $c_t$. The TCP-RM offset is $b_t$, and $e_t=\mathbf{1}\{r_t\notin I_t\}$ records an interval miss. Return values, interval endpoints and offsets use the same percentage-point scale.

\subsection{Split-conformal calibration}

Consider a score function $s(x,y)$ fitted using a separate training sample. A larger score indicates poorer agreement with the fitted model. Let $s_1,\ldots,s_m$ be its calibration scores and write their order statistics as $s_{(1)}\le\cdots\le s_{(m)}$. For a miscoverage level $\alpha$, define
\begin{equation}
 k=\left\lceil(m+1)(1-\alpha)\right\rceil,
 \qquad
 c=\begin{cases}
 s_{(k)},& k\le m,\\
 +\infty,& k=m+1.
 \end{cases}
 \label{eq:conformal-order}
\end{equation}
The prediction set is $\Gamma(x)=\{y:s(x,y)\le c\}$.

\begin{theorem}[Split-conformal coverage]
\label{thm:fs}
Assume the calibration and test scores are exchangeable conditional on the fitted score function. Then
\[
 \Pr\{Y_{\mathrm{new}}\in\Gamma(X_{\mathrm{new}})\}\ge 1-\alpha.
\]
\end{theorem}

This is the standard conformal rank guarantee \citep{Vovk2005}. CQR applies this argument to quantile-regression scores \citep{Romano2019}. The guarantee is marginal over the calibration and test observations. It does not provide coverage conditional on each feature value or on an arbitrary realised history. Ties can make the set conservative.

A rolling window does not establish exchangeability. In particular, temporal dependence and changes in the return distribution can invalidate the rank argument. We therefore assess coverage empirically for the financial series studied here. We do not claim finite-sample validity for these series from the use of a recent calibration window alone.

\subsection{Rolling TCP and its relation to CQR}

At forecast origin $t$, the preceding $w$ eligible observations are divided chronologically:
\[
 \mathcal{T}_t=\{t-w,\ldots,t-m-1\},\qquad
 \mathcal{C}_t=\{t-m,\ldots,t-1\}.
\]
We fit the lower and upper quantile functions, $\widehat q_{\alpha/2,t}$ and $\widehat q_{1-\alpha/2,t}$, on $\mathcal{T}_t$. Both fitted functions are then held fixed while scoring $\mathcal{C}_t$ and predicting $r_t$.

For each calibration observation, define
\begin{align}
 s_{i,t}&=\max\{\widehat q_{\alpha/2,t}(X_i)-r_i,\;
                 r_i-\widehat q_{1-\alpha/2,t}(X_i)\},\\
 s^+_{i,t}&=\max\{s_{i,t},0\}.
\end{align}
Rolling CQR uses $s_{i,t}$, which can be negative. TCP uses $s^+_{i,t}$. Let $c_t^{\mathrm{CQR}}$ and $c_t$ be their respective $k$th order statistics, with $k$ from \eqref{eq:conformal-order}. For $k\le m$,
\[
 c_t=\max\{c_t^{\mathrm{CQR}},0\}.
\]
TCP therefore leaves the base interval unchanged or expands it. CQR can also contract it. This truncation is the distinction between the two methods in our comparison; the training data and fitted quantiles are identical.

\begin{algorithm}[ht]
\caption{TCP forecast at observation $t$}
\label{alg:tcp}
\begin{algorithmic}[1]
\REQUIRE Past returns and features; window $w$; calibration size $m<w$; miscoverage $\alpha$ with $\lceil(m+1)(1-\alpha)\rceil\le m$
\STATE Set $\mathcal{T}_t=\{t-w,\ldots,t-m-1\}$ and $\mathcal{C}_t=\{t-m,\ldots,t-1\}$.
\STATE Fit $\widehat q_{\alpha/2,t}$ and $\widehat q_{1-\alpha/2,t}$ on $\{(X_i,r_i):i\in\mathcal{T}_t\}$.
\FOR{each $i\in\mathcal{C}_t$}
\STATE $s^+_{i,t}\leftarrow\max\{\widehat q_{\alpha/2,t}(X_i)-r_i,\;r_i-\widehat q_{1-\alpha/2,t}(X_i),\;0\}$.
\ENDFOR
\STATE Set $k=\lceil(m+1)(1-\alpha)\rceil$ and $c_t=s^+_{(k),t}$.
\STATE $\ell_t\leftarrow\widehat q_{\alpha/2,t}(X_t)-c_t$.
\STATE $u_t\leftarrow\widehat q_{1-\alpha/2,t}(X_t)+c_t$.
\ENSURE Interval $I_t=[\ell_t,u_t]$, issued before observing $r_t$.
\end{algorithmic}
\end{algorithm}

The numerical implementation rejects a requested finite threshold when $k>m$. It does not replace the unavailable order statistic with the largest calibration score. In the main two-sided experiment, $m=60$ and $\alpha=0.05$, so $k=58$.

\subsection{Online adjustment in TCP-RM}
\label{sec:adaptive}

TCP-RM uses the same rolling threshold $c_t$ and adds an offset $b_t$. Its interval is
\begin{equation}
 I_t^{\mathrm{RM}}=
 [\widehat q_{\alpha/2,t}(X_t)-(c_t+b_t),\;
  \widehat q_{1-\alpha/2,t}(X_t)+(c_t+b_t)].
 \label{eq:rm-interval}
\end{equation}
Let $t_0$ be the first forecast index and initialise $b_{t_0}=0$. After observing $r_t$, calculate $e_t=\mathbf{1}\{r_t\notin I_t^{\mathrm{RM}}\}$ and update
\begin{align}
 b_{t+1}&=\max\{b_t+\gamma_{j(t)}(e_t-\alpha),\;-c_t\},
 \label{eq:rm-update}\\
 \gamma_j&=\frac{\gamma_0}{(1+\lambda j)^\beta},
 \qquad j(t)=t-t_0+1.
 \label{eq:rm-stepsize}
\end{align}
The main settings are $\gamma_0=0.01$, $\lambda=0.1$ and $\beta=0.7$. The offset used in a forecast is determined before its outcome is observed.

The lower bound in \eqref{eq:rm-update} uses the current threshold $c_t$. It ensures $b_{t+1}\ge-c_t$. It does not ensure $c_{t+1}+b_{t+1}\ge0$, because the next rolling threshold can change. Thus TCP-RM can contract the next base interval. No additional projection onto $-c_{t+1}$ is applied.

Without the projection, the offset increment has the same coverage-error form as the quantile-tracking update of \citet{Angelopoulos2023}. They describe quantile tracking as P-control and discuss its accumulated-error interpretation. Their full PID construction also includes a separate error-integration rule. Decaying step sizes for online conformal prediction were studied by \citet{AngelopoulosBarberBates2024}. TCP-RM applies its offset to a newly computed rolling threshold and uses the projection in \eqref{eq:rm-update}. We do not claim that the recurrence or the use of a decaying step size is new. The guarantees for those earlier methods do not automatically apply to this implementation.

The step-size conditions $\sum_j\gamma_j=\infty$ and $\sum_j\gamma_j^2<\infty$ hold for the chosen decay. Those conditions alone do not establish calibration or convergence for the retrained forecaster, changing calibration threshold and projected offset used here. We make no asymptotic coverage or fixed-point claim for TCP-RM. Coverage backtests also do not establish exchangeability, mixing assumptions or the existence of a fixed calibration root.

\section{Methodology}
\label{sec:methodology}

\subsection{Data and forecast sample}

We analyse a fixed dataset of financial returns. It contains 1,720 dated observations from 10 November 2017 to 20 May 2025. The analysis covers 13 series: the S\&P 500, FTSE 100, Nikkei 225 and DAX indices; Bitcoin, Ether and Litecoin; gold, crude oil and natural gas futures; and the currency series labelled \texttt{EURUSD=X}, \texttt{GBPUSD=X} and \texttt{JPYUSD=X}. The S\&P 500, Bitcoin and gold are the three main examples. The remaining ten series assess whether their results extend to other assets. The dataset's VIX column is not used.

Return values are analysed on their stored percentage scale. We do not rescale the data before fitting. Interval widths are therefore measured in percentage points. The dataset does not include source prices or its construction script. We therefore cannot verify the return calculation or determine whether cryptocurrency returns across omitted dates were aggregated or sampled from daily returns.

All series share the dates in the return file. There are 437 gaps longer than one calendar day among the 1,719 successive date gaps. We retain these observations without adding missing dates. Grouping returns by the preceding calendar gap, Bitcoin's population variances for gaps of one to five days are 15.43, 21.04, 33.49, 42.36 and 61.94 squared percentage points. The corresponding S\&P 500 values are 1.55, 1.38, 2.26, 2.11 and 1.67. These patterns are consistent with different return horizons but do not establish how the file was constructed. Lags and rolling windows count recorded observations. The forecast target is therefore the next return in the file. A uniform one-calendar-day horizon is not established for every asset.

Feature construction removes the first 20 observations. A further 252 observations are required before the first forecast. Every method is evaluated on the same 1,448 dates, from 23 January 2019 to 20 May 2025. These dates and realised returns are checked for agreement across methods. The main comparison contains no missing forecasts.

\subsection{Features and rolling estimation}

For each asset, the feature vector contains five lagged returns, a volatility estimate, the squared previous return and its sign:
\[
 X_t=(r_{t-1},\ldots,r_{t-5},v_{20,t},r_{t-1}^2,
      \operatorname{sign}(r_{t-1})).
\]
The volatility feature is the sample standard deviation of the preceding 20 recorded returns:
\[
 \bar r_{20,t}=\frac1{20}\sum_{j=1}^{20}r_{t-j},\qquad
 v_{20,t}=\left\{\frac1{19}\sum_{j=1}^{20}
 (r_{t-j}-\bar r_{20,t})^2\right\}^{1/2}.
\]
Every feature is computed from past returns. No feature uses the return being predicted.

At each forecast origin, the 252 preceding eligible observations are split into 192 training observations and the most recent 60 calibration observations. A separate LightGBM quantile regressor is fitted for each required quantile \citep{Ke2017}. The main interval endpoints use levels 0.025 and 0.975. Lower-tail comparisons also fit levels 0.01 and 0.05 using the same training protocol.

We use LightGBM 4.7.0 with 100 boosting rounds, learning rate 0.1, at most 31 leaves per tree, and no explicit maximum-depth limit. The minimum leaf size is 20 observations and the minimum child Hessian sum is 0.001. The minimum split gain and both regularisation penalties are zero. All rows and features are available to each boosting iteration, with bagging disabled. The bin-construction subsample limit is 200,000. The random seed is zero, fitting uses one thread, and no early stopping or hyperparameter search is used. Models are refitted at every forecast origin. Predictions on the calibration and test observations are then shared across the conformal methods.

The quantile regressors are fitted separately. No sorting or rearrangement of their predicted endpoints is applied. We check endpoint order before interpreting widths and scores. None of the evaluated methods produces crossed intervals on the evaluation sample. This observation is not a general noncrossing guarantee. The analysis uses one fixed seed and does not estimate variability across seeds or alternative learner configurations.

\subsection{Forecasting methods}

\paragraph{Quantile regression and conformal methods.}
QR-base reports the uncalibrated 0.025 and 0.975 LightGBM predictions. TCP, TCP-RM and rolling CQR use those same predictions and the calibration rules in Section~\ref{sec:math}. Sharing the base predictions isolates the effect of the calibration rule.

\paragraph{Adaptive conformal inference.}
We implement a clipped, fixed-step version of ACI \citep{Gibbs2021}. It uses the same fitted quantiles and zero-truncated calibration scores as TCP. Initialise $a_{t_0}=\alpha$ and let
\[
 h_t=\operatorname{Quantile}_{\mathrm{higher}}
       (\{s^+_{i,t}:i\in\mathcal{C}_t\},1-a_t).
\]
The interval expands the base endpoints by $h_t$. After observing its error $e_t^{\mathrm{ACI}}$, update
\[
 a_{t+1}=\operatorname{clip}
 \{a_t+\eta(\alpha-e_t^{\mathrm{ACI}}),\;10^{-4},\;0.30\},
 \qquad \eta=0.05.
\]
For sorted scores of size $m$, the quantile convention above selects order statistic
$1+\lceil(m-1)(1-a_t)\rceil$. It differs from \eqref{eq:conformal-order}. The base quantile levels remain 0.025 and 0.975 as $a_t$ changes. The clipping and empirical-quantile convention are part of our implementation, so the original ACI guarantees are not asserted for this variant. We evaluate one step size. We do not compare aggregation across step sizes, as in AgACI \citep{zaffran2022adaptive}.

\paragraph{Fitted GARCH with Student-$t$ innovations.}
We fit a constant-mean GARCH(1,1) model \citep{Bollerslev1986},
\[
 r_t=\mu+\sigma_t z_t,\qquad
 \sigma_t^2=\omega+a_G(r_{t-1}-\mu)^2+b_G\sigma_{t-1}^2,
\]
where $z_t$ follows a Student-$t$ distribution scaled to unit variance. Parameters are re-estimated every 25 observations using at most the preceding 750 returns. The first estimation window contains 272 returns. Forecast quantiles are
\[
 \widehat q_{\tau,t}^{G}
 =\widehat\mu+\widehat\sigma_t
   \sqrt{\frac{\widehat\nu-2}{\widehat\nu}}\,
   t^{-1}_{\widehat\nu}(\tau),
\]
where $t^{-1}_{\nu}$ is the ordinary Student-$t$ quantile function.

Each refit uses four initialisations, with a fifth based on the preceding fit when available. We select the highest-likelihood candidate satisfying
$\omega>10^{-10}$, $a_G,b_G\ge10^{-6}$, $a_G+b_G\le1+10^{-6}$ and $\nu>2.05$.
If no candidate satisfies these checks, the best finite candidate with positive $\omega$ is used and flagged. If no usable fit is returned, the preceding parameters and their original estimation-window start are retained. The filter is not re-anchored to the failed refit's window. Near-unit persistence is permitted. Convergence, boundary flags and any carry-forward use are recorded.

Forecast variances are obtained by re-filtering the return history from the start of the selected estimation window. The initial filter variance is the population variance of its first 60 returns. The filter is updated between parameter refits. FHS and GARCH-EVT use this same fitted mean and variance filter.

\paragraph{Filtered historical simulation.}
FHS uses empirical quantiles of standardised residuals \citep{BaroneAdesi1999}. At each forecast origin, past returns in the current estimation window are filtered with the selected parameters. We retain at most the latest 500 finite standardised residuals. At least 100 are required to issue FHS and GARCH-EVT forecasts. Empirical quantiles use linear interpolation. They are multiplied by the forecast standard deviation and shifted by the fitted mean. All residuals precede the forecast date.

\paragraph{GARCH-EVT.}
Following the filtered extreme-value approach of \citet{McNeil2000}, we negate the standardised residuals and fit a generalised Pareto distribution to excess losses above their empirical 90th percentile. The location parameter of the excess distribution is fixed at zero. At least 25 excesses are required. Fits require a positive scale and a finite shape below one. Lower-tail quantiles are calculated from the fitted tail and mapped back to return units. If the requested tail probability is at least the observed exceedance fraction, the FHS quantile and expected shortfall are used instead. The upper 97.5\% endpoint is always the FHS endpoint. Thus the reported interval combines an EVT lower tail with an empirical upper tail.

The main evaluation needed no parameter carry-forward, no FHS substitution for an EVT lower-tail forecast, and no omission due to insufficient residuals or excesses. Of the 754 GARCH refits, 17 used the boundary fallback described above. All 754 selected fits reported optimiser convergence.

\paragraph{RiskMetrics and historical simulation.}
The RiskMetrics comparison uses a zero mean, Gaussian quantiles and the variance recursion
\[
 v_t=0.94v_{t-1}+0.06r_{t-1}^2
\]
\citep{Morgan1996}. Its initial variance is the population variance of the first 50 returns, all before the evaluation period. Historical simulation uses the empirical 0.025 and 0.975 quantiles of the preceding 252 returns, with linear interpolation.

\paragraph{Diagnostic controls.}
We include two diagnostic controls. The reversed-sign control uses the same scores, quantile rule, step size and clipping as the ACI comparator, but updates its level as
\[
 a_{t+1}=\operatorname{clip}
 \{a_t+\eta(e_t-\alpha),\;10^{-4},\;0.30\}.
\]
Here $e_t$ is this control's own interval-miss indicator. This update lowers the level after a covered observation and raises it after a miss. It is included to assess sensitivity to the update direction, not as an implementation of ACI.

The fixed-coefficient volatility control uses
\[
 v_t=10^{-6}+0.05r_{t-1}^2+0.9v_{t-1}
\]
with zero-mean Gaussian quantiles and the same initial-variance convention as RiskMetrics. Its parameters are never estimated. It assesses the behaviour of a fixed variance recursion and is distinct from the fitted GARCH comparison. The full interval comparison therefore contains ten forecasting methods and two diagnostic controls.

\subsection{Interval scores and paired comparisons}

For $T$ common forecast dates, empirical coverage and mean width are
\[
 \widehat{\mathrm{Cov}}=\frac1T\sum_t\mathbf{1}\{r_t\in[\ell_t,u_t]\},
 \qquad
 \overline W=\frac1T\sum_t(u_t-\ell_t).
\]
We also report the mean interval score \citep{Winkler1972}:
\begin{equation}
 S_{\alpha,t}=(u_t-\ell_t)
 +\frac{2}{\alpha}(\ell_t-r_t)\mathbf{1}\{r_t<\ell_t\}
 +\frac{2}{\alpha}(r_t-u_t)\mathbf{1}\{r_t>u_t\}.
 \label{eq:interval-score}
\end{equation}
Lower scores are preferred. This score penalises both width and missed observations. For a forecast lower quantile $q_{\tau,t}$, the pinball loss is
\[
 L_{\tau,t}=(r_t-q_{\tau,t})
 \{\tau-\mathbf{1}(r_t<q_{\tau,t})\}.
\]

We compare scores on matched dates using two-sided Diebold--Mariano tests \citep{DieboldMariano1995}. For each comparison, the loss difference is TCP's loss minus the competing method's loss. We use a Newey--West long-run variance estimate with Bartlett weights \citep{NeweyWest1987}. Our implementation chooses the lag truncation
$L=\lfloor4(T/100)^{2/9}\rfloor$, which gives seven lags for $T=1448$. The test uses a normal approximation. Reported significance uses a 5\% threshold without adjustment for multiple comparisons. A non-significant result is not evidence of equal predictive performance. These tests rely on conditions for inference from dependent loss sequences; the return data do not make those conditions automatic.

\subsection{Lower-tail forecasts and backtests}

We distinguish lower-tail quantile forecasts from the lower endpoints of two-sided prediction intervals. A jointly calibrated 95\% interval does not imply a separately calibrated 2.5\% lower tail. The TCP, TCP-RM, CQR and ACI lower endpoints are therefore labelled as 2.5\% diagnostics when tested against that reference level.

We also construct a separate one-sided conformal forecast, denoted CQR-1s. For a requested lower-tail level $\tau$, fit $\widehat q_{\tau,t}$ on the training slice and use the signed calibration scores
\[
 s^{\mathrm{lo}}_{i,t}=\widehat q_{\tau,t}(X_i)-r_i.
\]
With $k_\tau=\lceil(m+1)(1-\tau)\rceil$, its lower endpoint is
\[
 \ell^{\mathrm{1s}}_{\tau,t}
 =\widehat q_{\tau,t}(X_t)-s^{\mathrm{lo}}_{(k_\tau),t}.
\]
Under exchangeability, the order statistic gives a lower-tail miscoverage bound of
$1-k_\tau/(m+1)$. At $m=60$, the requested levels 0.025 and 0.05 correspond to reference bounds $1/61$ and $3/61$, respectively. We report backtests at the requested levels and separately at these attainable references. A 1\% request would require the unavailable 61st calibration order statistic. The implementation therefore omits that CQR-1s forecast. These rank-based bounds are not asserted as guarantees for the dependent return series.

For a lower endpoint $q_t$ and a stated reference probability $p$, an exception is
$H_t=\mathbf{1}\{r_t<q_t\}$. We report the exception count and rate, the unconditional-coverage test of \citet{Kupiec1995}, and the independence and conditional-coverage tests of \citet{Christoffersen1998}. The conditional-coverage statistic is the sum of the first two likelihood-ratio statistics. We also use the dynamic-quantile test of \citet{EngleManganelli2004}. Its regressors are a constant, four lags of $H_t-p$, and the current predicted lower endpoint.

All backtests state the probability used as their reference. A model is counted as passing a test when its $p$-value is strictly greater than 0.05. This convention does not establish correct specification. Upper-tail calibration is not tested separately.

\subsection{Expected shortfall and joint scores}

We compute expected shortfall (ES) for GARCH-t, FHS, GARCH-EVT, RiskMetrics, historical simulation and the fixed-coefficient volatility comparator. The requested lower-tail levels are 0.01, 0.025 and 0.05. ES uses the return sign convention: it describes the mean return in the lower tail, rather than a positive loss. Write its forecast as $\widehat{\mathrm{ES}}_{\tau,t}$, distinct from the interval-miss indicator $e_t$.

For GARCH-t, let $z_{\tau,\nu}=t_\nu^{-1}(\tau)$ and let $f_\nu$ be the ordinary Student-$t$ density. The forecast is
\[
 \widehat{\mathrm{ES}}^{G}_{\tau,t}
 =\widehat\mu-\widehat\sigma_t
 \sqrt{\frac{\widehat\nu-2}{\widehat\nu}}
 \frac{f_{\widehat\nu}(z_{\tau,\widehat\nu})}{\tau}
 \frac{\widehat\nu+z_{\tau,\widehat\nu}^{2}}{\widehat\nu-1}.
\]
FHS averages the standardised residuals at or below their empirical $\tau$ quantile. It then multiplies this mean by $\widehat\sigma_t$ and adds $\widehat\mu$.

GARCH-EVT uses the fitted Pareto tail for ES \citep{McNeil2000}. Let $u_t^{\mathrm{EVT}}$ be the loss threshold, $\xi_t$ the fitted shape, $d_t$ the fitted scale, and $v_{\tau,t}$ the fitted loss quantile on the standardised scale. Then
\[
 \widehat{\mathrm{ES}}^{\mathrm{EVT}}_{\tau,t}
 =\widehat\mu-\widehat\sigma_t
 \frac{v_{\tau,t}+d_t-\xi_t u_t^{\mathrm{EVT}}}{1-\xi_t}.
\]
The FHS substitution described above also applies to ES when the requested probability lies outside the fitted tail.

RiskMetrics and the fixed-coefficient comparator use the Gaussian formula
\[
 \widehat{\mathrm{ES}}_{\tau,t}
 =-\sigma_t\frac{\phi(\Phi^{-1}(\tau))}{\tau},
\]
where $\phi$ and $\Phi$ are the standard normal density and distribution function. Historical simulation averages returns at or below the empirical $\tau$ quantile in its preceding 252-observation window. The empirical tail-mean calculations return the quantile if their selected tail is empty.

For a quantile forecast $q_{\tau,t}$ and an ES forecast $E_{\tau,t}<0$, we use the FZ0 joint loss of \citet{Patton2019}:
\begin{equation}
 \begin{split}
 S^{\mathrm{FZ0}}_{\tau,t}
 ={}&-\frac{\mathbf{1}\{r_t\le q_{\tau,t}\}
                  (q_{\tau,t}-r_t)}{\tau E_{\tau,t}}
      +\frac{q_{\tau,t}}{E_{\tau,t}}\\
    &+\log(-E_{\tau,t})-1,
 \qquad E_{\tau,t}=\widehat{\mathrm{ES}}_{\tau,t}.
 \end{split}
 \label{eq:fz0}
\end{equation}
We report its mean, with lower values preferred. Comparisons use the same asset, return scale and tail level. The implementation excludes ES values at or above $-10^{-12}$ from this calculation.

The output column \texttt{ES\_ratio} is a separate descriptive diagnostic. For
$A_\tau=\{t:r_t<q_{\tau,t}\}$, it is
\[
 \mathrm{ES\_ratio}_\tau
 =\frac{|A_\tau|^{-1}\sum_{t\in A_\tau}r_t}
        {|A_\tau|^{-1}\sum_{t\in A_\tau}
                         \widehat{\mathrm{ES}}_{\tau,t}}.
\]
The ratio is omitted when there are fewer than five exceptions. A value near one indicates agreement between these two averages; it is not a formal ES backtest.

The conformal methods do not produce ES forecasts in this study. Their ES and FZ0 entries are therefore unavailable. We make no claim about TCP's or TCP-RM's expected-shortfall accuracy or joint VaR/ES performance.

\subsection{Computational settings}

The main forecast calculations use Python 3.11 on Windows, with NumPy 2.4.4, pandas 3.0.2, SciPy 1.17.1, LightGBM 4.7.0 and \texttt{arch} 8.0.0. GARCH optimisation allows up to 800 iterations per initialisation. Numerical libraries are restricted to one thread.

Fitted GARCH solutions can depend on numerical initialisation and the computing environment. This can affect whether a boundary fallback is selected. Backtest decisions close to the 5\% threshold should therefore be interpreted with care.

\section{Results and Discussion}
\label{sec:result}

We compare ten forecasting methods and two diagnostic controls on thirteen return series. Every method uses the same 1,448 forecast dates, from 23 January 2019 to 20 May 2025, with a nominal 95\% coverage target. We report coverage, mean width and the Winkler interval score, which penalises both width and misses.

\subsection{Full-sample interval performance}

Table~\ref{tab:results} reports the S\&P 500, Bitcoin and Gold results. TCP attains coverage of 0.9523, 0.9537 and 0.9427, respectively. Its intervals are wider than those of historical simulation on all three series. QR-base produces narrower intervals, but its coverage falls to 0.8377, 0.8550 and 0.8101. Clipped ACI attains coverage of 0.9441, 0.9448 and 0.9399. The two diagnostic controls are reported separately from these comparisons and do not represent correctly specified ACI or fitted GARCH models.

Across the thirteen series, CQR has the smallest mean absolute deviation from the 0.95 coverage target, at 0.003272. TCP follows at 0.003432, then TCP-RM at 0.003485. Appendix~\ref{app:all-series} reports the coverage deviations, mean interval-score ranks and paired-test counts across all thirteen series. These are descriptive comparisons of full-sample coverage. They do not establish conditional coverage or performance during a particular market event.

TCP and CQR are closely related in this experiment. Their intervals differ on 688 of the 18,824 forecasts, or 3.655\%. TCP is never narrower because its calibration threshold is truncated at zero. The Diebold--Mariano tests find no significant interval-score difference between TCP and CQR on any of the thirteen series at the 5\% level. This does not establish equivalence.

Coverage and interval score give different rankings. Against GARCH-t, FHS, GARCH-EVT and RiskMetrics, TCP has a significantly higher mean interval score in 42 of 52 comparisons. The remaining ten show no significant difference. None favours TCP at the 5\% level. These are individual tests without a multiple-testing adjustment. The raw scores also need a qualification: on EURUSD, TCP's score of 2.5332 is lower than FHS's 2.5443 and GARCH-EVT's 2.5379. Neither difference is significant. Thus, closeness to the coverage target does not translate into a general interval-score advantage for TCP.

\begin{table}[p]
\centering
\small
\caption{Interval forecasts for the three main assets. All rows use the same 1,448 forecast dates and a 95\% coverage target. Width and the Winkler interval score are in percentage points. The last two rows for each asset are diagnostic controls. Lower interval scores are preferred.}
\label{tab:results}
\begin{tabular}{lrrr}
\toprule
Method & Coverage & Mean width & Winkler score \\
\midrule
\multicolumn{4}{l}{\textit{S\&P 500}} \\
TCP & 0.9523 & 5.2097 & 7.1168 \\
TCP-RM & 0.9530 & 5.2171 & 7.1157 \\
CQR & 0.9510 & 5.1761 & 7.1149 \\
ACI (clipped) & 0.9441 & 5.2163 & 6.9431 \\
QR-base & 0.8377 & 3.1609 & 9.7240 \\
GARCH-t & 0.9496 & 4.6068 & 6.1435 \\
FHS & 0.9468 & 4.4940 & 5.9937 \\
GARCH-EVT & 0.9482 & 4.5285 & 6.0054 \\
RiskMetrics & 0.9434 & 4.4752 & 6.4034 \\
Historical simulation & 0.9316 & 5.0705 & 8.6679 \\
\addlinespace[3pt]
Reversed-sign control & 0.8736 & 3.7894 & 9.2980 \\
Fixed-coefficient control & 0.8267 & 3.1012 & 7.4492 \\
\midrule
\multicolumn{4}{l}{\textit{Bitcoin}} \\
TCP & 0.9537 & 20.8879 & 27.9846 \\
TCP-RM & 0.9537 & 20.8859 & 27.9851 \\
CQR & 0.9537 & 20.8815 & 28.0075 \\
ACI (clipped) & 0.9448 & 21.1456 & 28.7839 \\
QR-base & 0.8550 & 12.7594 & 31.9108 \\
GARCH-t & 0.9413 & 16.6101 & 26.1980 \\
FHS & 0.9448 & 17.9364 & 26.3508 \\
GARCH-EVT & 0.9454 & 18.1520 & 26.3508 \\
RiskMetrics & 0.9344 & 16.1650 & 26.3141 \\
Historical simulation & 0.9434 & 17.9999 & 26.5294 \\
\addlinespace[3pt]
Reversed-sign control & 0.8874 & 16.3529 & 30.5044 \\
Fixed-coefficient control & 0.8543 & 11.1649 & 30.1409 \\
\midrule
\multicolumn{4}{l}{\textit{Gold}} \\
TCP & 0.9427 & 4.7368 & 6.3246 \\
TCP-RM & 0.9427 & 4.7438 & 6.3246 \\
CQR & 0.9427 & 4.7352 & 6.3252 \\
ACI (clipped) & 0.9399 & 4.9225 & 6.3765 \\
QR-base & 0.8101 & 2.8179 & 8.2265 \\
GARCH-t & 0.9289 & 3.9401 & 6.0405 \\
FHS & 0.9358 & 4.0454 & 6.0071 \\
GARCH-EVT & 0.9385 & 4.1078 & 5.9842 \\
RiskMetrics & 0.9247 & 3.8885 & 5.8432 \\
Historical simulation & 0.9323 & 4.0409 & 6.1032 \\
\addlinespace[3pt]
Reversed-sign control & 0.8294 & 2.9901 & 7.9719 \\
Fixed-coefficient control & 0.8363 & 2.7309 & 7.1495 \\
\bottomrule
\end{tabular}
\end{table}

\subsection{Effect of the online offset}

At the main settings, TCP-RM changes little relative to TCP. On the S\&P 500, coverage rises from 0.9523 to 0.9530, which is one additional covered observation. Coverage is unchanged for Bitcoin and Gold. The absolute changes in mean width are below 0.008 percentage points on each of these three series. The S\&P 500 mean widths are 5.2097 for TCP and 5.2171 for TCP-RM.

Across all thirteen series, the largest absolute realised offset is 0.016625 percentage points. The interval-score tests find no significant TCP--TCP-RM difference on twelve series. The remaining comparison, EURUSD, favours TCP, although its mean score advantage is only 0.000663. There is no significant comparison favouring TCP-RM. Appendix~\ref{app:traces} shows the full-sample intervals, rolling coverage and offsets for the three main assets.

The expansion $c_t+b_t$ is negative on 61 forecasts across twelve series, so TCP-RM contracts the base interval on those dates. None of the evaluated intervals has crossed endpoints. As Appendix~\ref{app:rm-properties} explains, the projection does not guarantee a nonnegative expansion at the next forecast origin.

\subsection{Performance during the COVID window}

\begin{figure}[p]
\centering
\includegraphics[width=\linewidth]{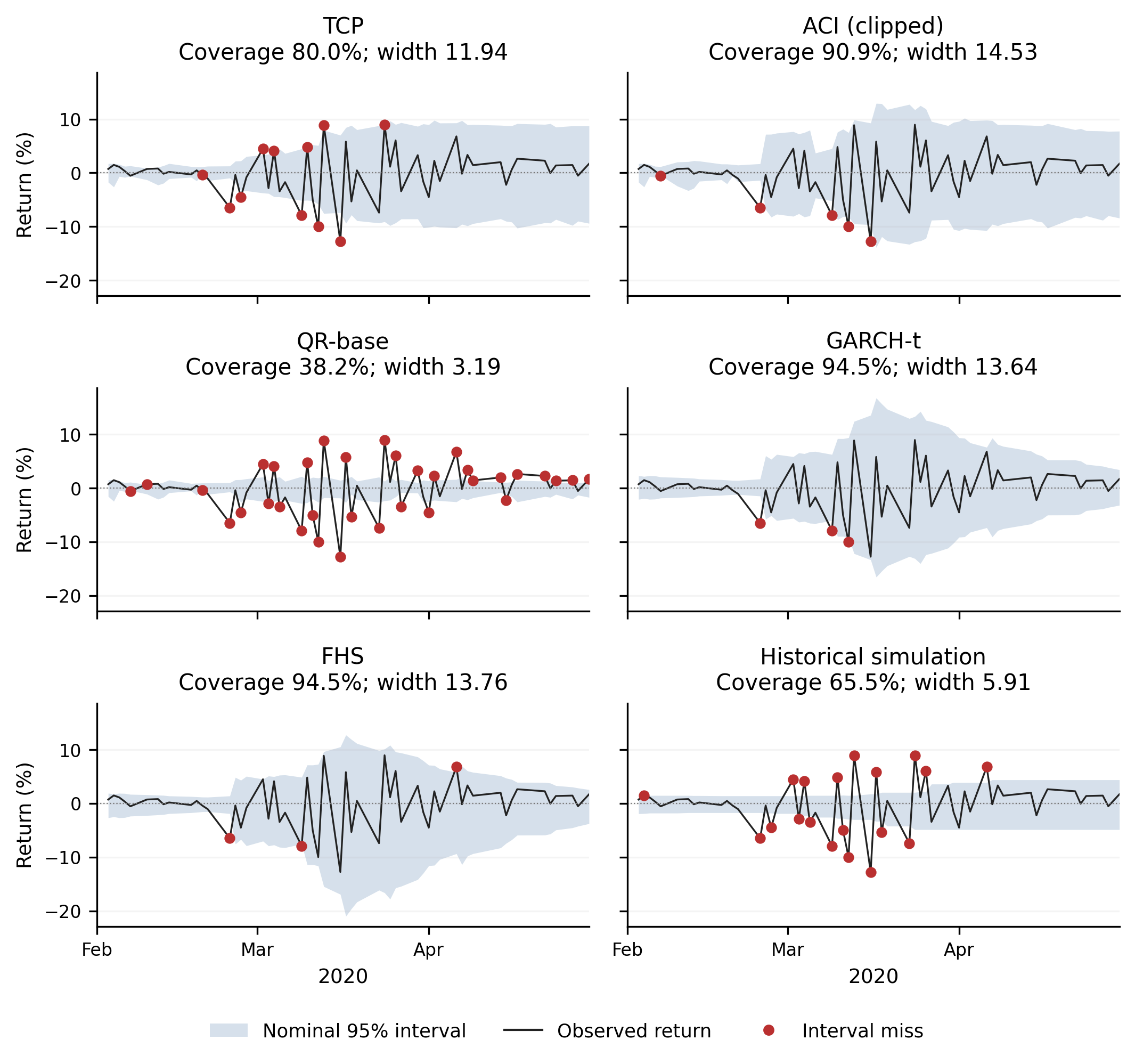}
\caption{S\&P 500 prediction intervals from February to April 2020, with a nominal 95\% coverage target. Blue bands show the intervals, black lines show observed returns, and red points mark interval misses. All six panels use the same dates and vertical scale. Panel titles report coverage and mean width over the 55 recorded observations. Returns and widths are in percentage points.}
\label{fig:viz}
\end{figure}

The period from February to April 2020 contains 55 recorded observations for each series. Table~\ref{tab:covwidth-sp500-crisis} reports S\&P 500 performance in this window. TCP and TCP-RM both cover 80\% of observations. GARCH-t, FHS and GARCH-EVT each cover 94.55\%, while clipped ACI covers 90.91\%. FHS has a mean interval score of 18.0868, compared with TCP's 29.0316. Figure~\ref{fig:viz} shows the intervals and missed observations for six methods.

The ordering differs across assets. On Gold, TCP covers 87.27\% of observations in this window, compared with 70.91\% for GARCH-t and 80\% for FHS. Clipped ACI covers 90.91\%. The Bitcoin and Gold tables and interval plots are in Appendix~\ref{app:covwidth-crisis}. These short-window results do not support a single ranking across markets. They also show why full-sample coverage near 95\% should not be read as protection against undercoverage during a crisis.

\begin{table}[tbp]
\centering
\small
\caption{S\&P 500 interval performance from February to April 2020. Each method is evaluated on the same 55 recorded observations. The coverage target is 95\%. Width and the Winkler score are in percentage points.}
\label{tab:covwidth-sp500-crisis}
\begin{tabular}{lrrr}
\toprule
Method & Coverage & Mean width & Winkler score \\
\midrule
TCP & 0.8000 & 11.9371 & 29.0316 \\
TCP-RM & 0.8000 & 11.9511 & 29.0001 \\
CQR & 0.8000 & 11.9371 & 29.0316 \\
ACI (clipped) & 0.9091 & 14.5301 & 23.7701 \\
QR-base & 0.3818 & 3.1894 & 73.7764 \\
GARCH-t & 0.9455 & 13.6409 & 19.2597 \\
FHS & 0.9455 & 13.7554 & 18.0868 \\
GARCH-EVT & 0.9455 & 13.7177 & 18.1273 \\
RiskMetrics & 0.8545 & 12.7000 & 26.2580 \\
Historical simulation & 0.6545 & 5.9058 & 59.3025 \\
\bottomrule
\end{tabular}
\end{table}

\begin{figure}[p]
\centering
\includegraphics[width=\linewidth]{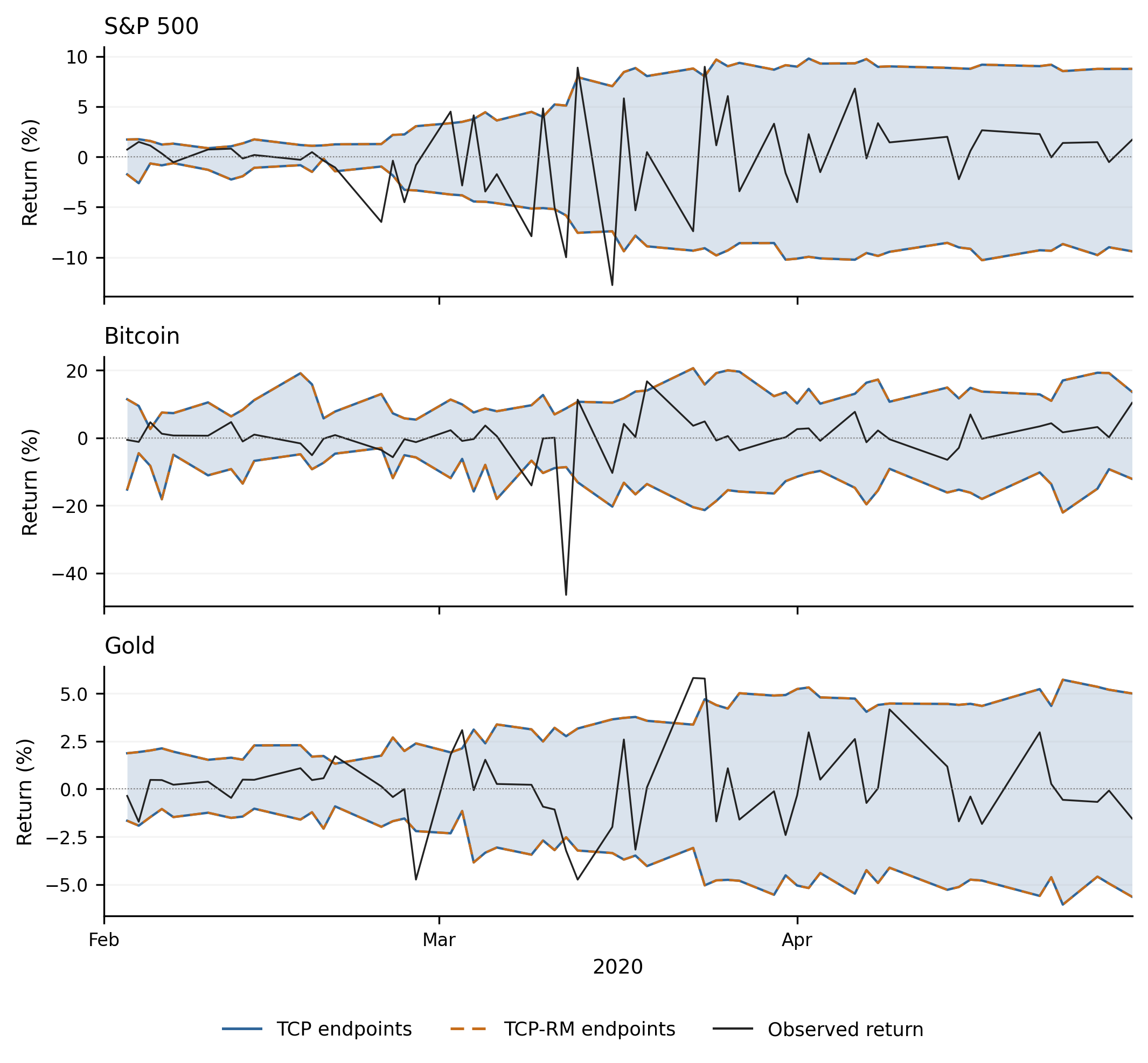}
\caption{TCP and TCP-RM intervals from February to April 2020 for the S\&P 500, Bitcoin and Gold. Solid blue lines mark TCP endpoints and dashed orange lines mark TCP-RM endpoints. The shaded region is the TCP interval. Black lines show observed returns. The two methods' endpoints nearly coincide at the main settings. Each panel uses its own vertical scale.}
\label{fig:tcp-tcprm-sp500}
\end{figure}

Figure~\ref{fig:tcp-tcprm-sp500} compares TCP and TCP-RM on the three main assets during the same window. Their endpoints nearly coincide. Their coverage is identical within each asset: 0.8000 for the S\&P 500, 0.8909 for Bitcoin and 0.8727 for Gold. Mean widths are 11.9371 versus 11.9511, 24.2814 versus 24.2729, and 6.8087 versus 6.8119, respectively.

\subsection{Backtests of lower-tail coverage}

Table~\ref{tab:backtests-sp500} tests lower-tail exception sequences. For TCP, TCP-RM, CQR and the ACI constructions, these are diagnostics of the lower endpoint of a two-sided interval. A 95\% two-sided coverage target does not impose a 2.5\% exception probability on each tail. CQR-1s instead constructs a one-sided forecast. We distinguish these uses when interpreting the tests.

On the S\&P 500, the TCP lower endpoint has 49 exceptions, rather than the 36.2 expected at a 2.5\% reference probability. Its unconditional-coverage and dynamic-quantile tests reject at the 5\% level. TCP-RM has 48 exceptions and fails only the dynamic-quantile test. FHS and GARCH-EVT have 36 and 34 exceptions, respectively. Both have unconditional-coverage, conditional-coverage and dynamic-quantile $p$-values greater than 0.05 on this series.

Across the thirteen series, those three $p$-values all exceed 0.05 on one series for the TCP lower-endpoint diagnostic, twelve for FHS, eleven for GARCH-EVT and seven for GARCH-t. These counts can change with small numerical differences in volatility-model fits, especially near the 5\% threshold. Passing the tests does not prove calibration. In particular, FHS and GARCH-EVT still fail the Gold dynamic-quantile test, with $p$-values of 0.000396 and 0.000696.

With 60 calibration observations, CQR-1s at requested level $\tau=0.025$ uses the attainable rank level $1/61$. On the S\&P 500, it produces 31 exceptions. Table~\ref{tab:backtests-sp500} reports tests against both reference probabilities for the same forecast sequence. At $1/61$, unconditional-coverage and conditional-coverage $p$-values exceed 0.05 on all thirteen series, but the dynamic-quantile test does so on only two. Appendix~\ref{app:backtests} gives the Bitcoin and Gold backtests and the test counts across all thirteen series. The difference between these tests matters: an acceptable overall exception rate does not establish temporal calibration.

\begin{table}[tbp]
\centering
\small
\setlength{\tabcolsep}{4pt}
\caption{S\&P 500 lower-tail backtests on 1,448 observations. $p$ is the reference exception probability; the remaining $p$-columns contain test $p$-values. A dagger marks a two-sided interval endpoint tested as a 2.5\% diagnostic. The two CQR-1s rows use the same forecasts, constructed at requested level $\tau=0.025$, and test them at the requested and attainable reference levels. UC, IND, CC and DQ denote unconditional coverage, independence, conditional coverage and dynamic quantile tests.}
\label{tab:backtests-sp500}
\begin{tabular}{lrrrrrr}
\toprule
Forecast & $p$ & Exceptions & $p_{\rm UC}$ & $p_{\rm IND}$ & $p_{\rm CC}$ & $p_{\rm DQ}$ \\
\midrule
TCP$^\dagger$ & 0.025 & 49 & 0.0407 & 0.3297 & 0.0767 & 0.0008 \\
TCP-RM$^\dagger$ & 0.025 & 48 & 0.0583 & 0.3003 & 0.0974 & 0.0008 \\
CQR$^\dagger$ & 0.025 & 50 & 0.0279 & 0.3608 & 0.0588 & 0.0007 \\
ACI (clipped)$^\dagger$ & 0.025 & 55 & 0.0032 & 0.3859 & 0.0090 & 0.0001 \\
QR-base & 0.025 & 120 & $<10^{-4}$ & 0.7388 & $<10^{-4}$ & $<10^{-4}$ \\
GARCH-t & 0.025 & 56 & 0.0020 & 0.9048 & 0.0084 & 0.0121 \\
FHS & 0.025 & 36 & 0.9731 & 0.9115 & 0.9933 & 0.8552 \\
GARCH-EVT & 0.025 & 34 & 0.7084 & 0.8243 & 0.9097 & 0.7070 \\
RiskMetrics & 0.025 & 57 & 0.0012 & 0.8623 & 0.0053 & 0.0003 \\
Historical simulation & 0.025 & 50 & 0.0279 & 0.0330 & 0.0092 & $<10^{-4}$ \\
Reversed-sign control$^\dagger$ & 0.025 & 103 & $<10^{-4}$ & 0.5197 & $<10^{-4}$ & $<10^{-4}$ \\
Fixed-coefficient control & 0.025 & 124 & $<10^{-4}$ & 0.4388 & $<10^{-4}$ & $<10^{-4}$ \\
\midrule
CQR-1s & 0.025 & 31 & 0.3697 & 0.6944 & 0.6191 & 0.4089 \\
CQR-1s & $1/61$ & 31 & 0.1510 & 0.6944 & 0.3302 & 0.1068 \\
\bottomrule
\end{tabular}
\end{table}

\subsection{Sensitivity analysis}

We vary the total window $w\in\{100,252,500\}$, calibration size $m\in\{40,60\}$ and initial step size $\gamma_0\in\{0.005,0.01,0.05\}$. We hold $\lambda=0.1$ and $\beta=0.7$ fixed. Each window and calibration setting also includes TCP without the online offset. The training sample contains $w-m$ observations. Table~\ref{tab:sensitivity-sp500} reports the S\&P 500 results, including the main setting $w=252$, $m=60$ and $\gamma_0=0.01$.

At $w=252$ and $m=60$, increasing $\gamma_0$ from 0.005 to 0.05 raises mean width from 5.2134 to 5.2421. Coverage rises from 0.9530 to 0.9544, while the interval score changes from 7.1162 to 7.1149. Larger steps do not consistently improve the score. At $w=500$ and $m=60$, the same increase raises width from 5.4028 to 5.6095 and the score from 7.3404 to 7.3651.

The window comparisons use different evaluation periods. Their sample sizes are 1,600, 1,448 and 1,200 for $w=100$, 252 and 500, respectively. Changing $m$ also changes the training size and the available calibration ranks. The table therefore does not isolate a window-length effect or justify an optimal setting. Comparisons across step sizes within a fixed $(w,m)$ pair use the same dates and calibration size. Appendix~\ref{app:sensitivity} reports the corresponding Bitcoin and Gold results, including TCP without the offset.

\begin{table}[p]
\centering
\small
\caption{S\&P 500 sensitivity to the total window $w$, calibration size $m$ and initial step size $\gamma_0$. A dash denotes TCP without the online offset. All other rows use TCP-RM. The training size is $w-m$. Sample length $n$ changes with $w$. Width and the Winkler score are in percentage points.}
\label{tab:sensitivity-sp500}
\begin{tabular}{rrrrrrr}
\toprule
$w$ & $m$ & $\gamma_0$ & $n$ & Coverage & Mean width & Winkler score \\
\midrule
100 & 40 & -- & 1600 & 0.9575 & 5.3038 & 6.8462 \\
100 & 40 & 0.005 & 1600 & 0.9575 & 5.3052 & 6.8465 \\
100 & 40 & 0.010 & 1600 & 0.9575 & 5.3066 & 6.8468 \\
100 & 40 & 0.050 & 1600 & 0.9581 & 5.3172 & 6.8493 \\
\midrule
100 & 60 & -- & 1600 & 0.9513 & 5.5086 & 7.4335 \\
100 & 60 & 0.005 & 1600 & 0.9513 & 5.5153 & 7.4339 \\
100 & 60 & 0.010 & 1600 & 0.9513 & 5.5220 & 7.4343 \\
100 & 60 & 0.050 & 1600 & 0.9525 & 5.5547 & 7.4371 \\
\midrule
252 & 40 & -- & 1448 & 0.9572 & 5.1145 & 6.7395 \\
252 & 40 & 0.005 & 1448 & 0.9579 & 5.1168 & 6.7394 \\
252 & 40 & 0.010 & 1448 & 0.9579 & 5.1191 & 6.7393 \\
252 & 40 & 0.050 & 1448 & 0.9586 & 5.1314 & 6.7380 \\
\midrule
252 & 60 & -- & 1448 & 0.9523 & 5.2097 & 7.1168 \\
252 & 60 & 0.005 & 1448 & 0.9530 & 5.2134 & 7.1162 \\
252 & 60 & 0.010 & 1448 & 0.9530 & 5.2171 & 7.1157 \\
252 & 60 & 0.050 & 1448 & 0.9544 & 5.2421 & 7.1149 \\
\midrule
500 & 40 & -- & 1200 & 0.9508 & 5.3951 & 7.0995 \\
500 & 40 & 0.005 & 1200 & 0.9525 & 5.4135 & 7.1013 \\
500 & 40 & 0.010 & 1200 & 0.9525 & 5.4319 & 7.1033 \\
500 & 40 & 0.050 & 1200 & 0.9567 & 5.5348 & 7.1195 \\
\midrule
500 & 60 & -- & 1200 & 0.9483 & 5.3755 & 7.3394 \\
500 & 60 & 0.005 & 1200 & 0.9492 & 5.4028 & 7.3404 \\
500 & 60 & 0.010 & 1200 & 0.9508 & 5.4297 & 7.3419 \\
500 & 60 & 0.050 & 1200 & 0.9567 & 5.6095 & 7.3651 \\
\bottomrule
\end{tabular}
\end{table}

These comparisons use one base learner, one seed and the stated grid. Clipped ACI uses one fixed step size, without adaptive aggregation over step sizes.


\section{Conclusion and Future Work}
\label{sec:conclusion}

We studied rolling conformal prediction intervals for financial returns. TCP uses the CQR calibration score truncated at zero. For TCP, this prevents contraction of the base interval. TCP-RM adds an online offset based on past coverage errors. Its total expansion can be negative, so TCP-RM can still contract the base interval. These constructions use established conformal calibration and quantile-tracking ideas.

The comparison covered thirteen series on the same 1,448 forecast dates. TCP's full-sample coverage was close to the 95\% target. CQR had the smallest mean absolute deviation from that target, followed by TCP and TCP-RM. Coverage and interval scores gave different rankings. Of the 52 interval-score comparisons between TCP and the four volatility models, 42 favoured the volatility model at the unadjusted 5\% significance level. The remaining ten showed no significant difference. These results do not establish a general forecasting advantage for TCP.

At the main settings, TCP-RM changed coverage and mean width very little. Interval-score tests found no significant advantage over TCP on any series.

Full-sample coverage also concealed periods of undercoverage. During the February--April 2020 window, TCP and TCP-RM each covered 80\% of the recorded S\&P 500 returns. On Gold in the same window, TCP covered 87.27\%, compared with 70.91\% for GARCH-t. These short-window results do not establish a single ranking across markets. Coverage near the target over the full sample did not ensure coverage near the target during each market event.

Two-sided interval coverage must also be distinguished from lower-tail calibration. TCP's lower endpoint, evaluated as a 2.5\% diagnostic, passed the unconditional-coverage, conditional-coverage and dynamic-quantile tests together on only one of thirteen series. For CQR-1s, evaluation at the attainable $1/61$ reference gave unconditional-coverage and conditional-coverage passes on all thirteen series. The dynamic-quantile test passed on only two. An acceptable overall exception rate does not establish temporal calibration. The conformal methods do not provide expected-shortfall forecasts in this study.

The finite-sample rank guarantee requires exchangeable calibration and test scores. A rolling window does not ensure this condition for financial returns. We have not established asymptotic calibration for TCP-RM. The empirical comparison also uses one base learner, one random seed and one fixed step size for clipped ACI. The window sensitivity results use different evaluation periods, so they cannot isolate the effect of window length.

Future work should compare base forecasters and adaptive calibration rules on common evaluation dates. Step sizes and window lengths should be selected using data separate from the final evaluation sample. Coverage within volatility regimes would help assess failures that full-sample averages can conceal. Further tail-risk studies should evaluate one-sided quantile and expected-shortfall forecasts directly. Multivariate and portfolio extensions would require their own calibration arguments and empirical evaluation.


\section*{Code and Data Availability}
The code and study dataset, including the scripts used to
generate the tables and figures, are available at
\url{https://github.com/agnivibes/temporal-conformal-prediction-tcp}.


\bibliographystyle{plainnat}
\bibliography{references}


\appendix
\section{Mathematical details}
\label{app:math}

\subsection{Proof of the split-conformal rank guarantee}
\label{app:rank-proof}

We prove Theorem~\ref{thm:fs} under its stated exchangeability assumption. This is the standard conformal rank argument \citep{Vovk2005}. It does not assume that the financial observations in this study satisfy that assumption.

\paragraph{Proof.}
Condition on the fitted score function. Write the calibration scores as $S_1,\ldots,S_m$ and the test score as $S_{m+1}$. Introduce independent continuous uniform variables $U_1,\ldots,U_{m+1}$, also independent of the scores. Order the pairs $(S_i,U_i)$ first by score and then by $U_i$ to break ties. The pairs remain exchangeable and have distinct ranks with probability one. The test pair's rank $R$ is therefore uniform on $\{1,\ldots,m+1\}$, conditional on the fitted score function.

Let $k=\lceil(m+1)(1-\alpha)\rceil$. If $k\le m$, denote the $k$th smallest calibration score by $S^{\mathrm{cal}}_{(k)}$. Whenever $R\le k$, at most $k-1$ calibration pairs precede the test pair. It follows that
\[
 S_{m+1}\le S^{\mathrm{cal}}_{(k)}.
\]
Thus
\[
 \Pr\{S_{m+1}\le S^{\mathrm{cal}}_{(k)}
       \mid\text{fitted score function}\}
 \ge \Pr\{R\le k\mid\text{fitted score function}\}
 =\frac{k}{m+1}\ge 1-\alpha.
\]
If $k=m+1$, the threshold is $+\infty$ and the set covers every finite test score. Averaging over the fitted score function gives the stated marginal bound. The auxiliary uniforms are used only in the proof; the implemented prediction set includes scores tied at its threshold. Ties can therefore increase coverage.\hfill$\square$

\subsection{Step sizes and finite-horizon bounds for TCP-RM}
\label{app:rm-properties}

For $\gamma_0,\lambda>0$, the step size in \eqref{eq:rm-stepsize} satisfies
\[
 \gamma_j\sim \gamma_0\lambda^{-\beta}j^{-\beta}
 \quad\text{as }j\to\infty.
\]
Consequently, $\sum_j\gamma_j=\infty$ and $\sum_j\gamma_j^2<\infty$ when $1/2<\beta\le1$. The choice $\beta=0.7$ meets both conditions. These series calculations do not establish calibration or convergence of the changing forecaster.

There is also a deterministic bound on the realised offset. Starting from $b_{t_0}=0$, after $J$ updates,
\begin{equation}
 -\alpha\sum_{j=1}^{J}\gamma_j
 \ \le\ b_{t_0+J}\ \le\
 (1-\alpha)\sum_{j=1}^{J}\gamma_j.
 \label{eq:offset-bounds}
\end{equation}
For the lower bound, each unprojected update decreases the offset by at most $\alpha\gamma_j$, and the projection can only raise it. For the upper bound, each unprojected update increases it by at most $(1-\alpha)\gamma_j$. The projection value $-c_t$ is nonpositive because $c_t\ge0$, so it cannot exceed the nonnegative upper bound. Induction gives \eqref{eq:offset-bounds}.

At the main settings, the sum of the first 1,448 step sizes is 1.147834. The bounds after all 1,448 updates are therefore $-0.057392$ and $1.090442$ percentage points. The final such update is made after the last evaluated forecast. A forecast issued earlier uses only the preceding updates. These bounds limit the possible offset movement; they do not guarantee a particular coverage rate.

The projection gives $b_{t+1}\ge-c_t$, not $b_{t+1}\ge-c_{t+1}$. If the next calibration threshold falls sufficiently, $c_{t+1}+b_{t+1}$ can be negative. None of these calculations establishes a fixed calibration root or asymptotic calibration for TCP-RM.

\clearpage
\section{Interval results across thirteen series}
\label{app:all-series}

Table~\ref{tab:all-series-summary} summarises the common evaluation sample. Coverage MAD measures closeness to the 0.95 target, whereas the interval-score rank also reflects interval width and the sizes of misses. A lower value is preferred for both summaries. Ranks compare methods within a series before averaging; raw interval scores are not averaged across assets with different return scales.

The test counts compare TCP with each listed method on all thirteen series. The counts without a significant difference do not establish equal performance. The two controls remain in the rank calculation and are identified separately in the table.

\begin{table}[htbp]
\centering
\small
\setlength{\tabcolsep}{4pt}
\caption{Interval results across thirteen series. Coverage MAD is the mean absolute difference between each series' coverage and 0.95. Interval-score ranks are calculated within each series across all twelve rows, with rank one best, and then averaged. The last three columns count two-sided Diebold--Mariano comparisons of TCP against the listed method at the unadjusted 5\% level. ND denotes no significant difference.}
\label{tab:all-series-summary}
\begin{tabular}{lrrrrr}
\toprule
Method & Coverage MAD & Mean rank & TCP better & ND & Other better \\
\midrule
TCP & 0.003432 & 6.462 & -- & -- & -- \\
TCP-RM & 0.003485 & 6.462 & 1 & 12 & 0 \\
CQR & 0.003272 & 7.231 & 0 & 13 & 0 \\
ACI (clipped) & 0.006045 & 7.000 & 2 & 11 & 0 \\
QR-base & 0.116702 & 11.846 & 13 & 0 & 0 \\
GARCH-t & 0.005450 & 2.154 & 0 & 2 & 11 \\
FHS & 0.006024 & 3.077 & 0 & 3 & 10 \\
GARCH-EVT & 0.005206 & 2.385 & 0 & 3 & 10 \\
RiskMetrics & 0.014705 & 2.769 & 0 & 2 & 11 \\
Historical simulation & 0.014811 & 7.692 & 2 & 9 & 2 \\
\midrule
Reversed-sign control & 0.095187 & 11.154 & 13 & 0 & 0 \\
Fixed-coefficient control & 0.112718 & 9.769 & 7 & 6 & 0 \\
\bottomrule
\end{tabular}
\end{table}

\clearpage
\section{Additional crisis-window results}
\label{app:covwidth-crisis}

\begin{figure}[p]
\centering
\includegraphics[width=\linewidth]{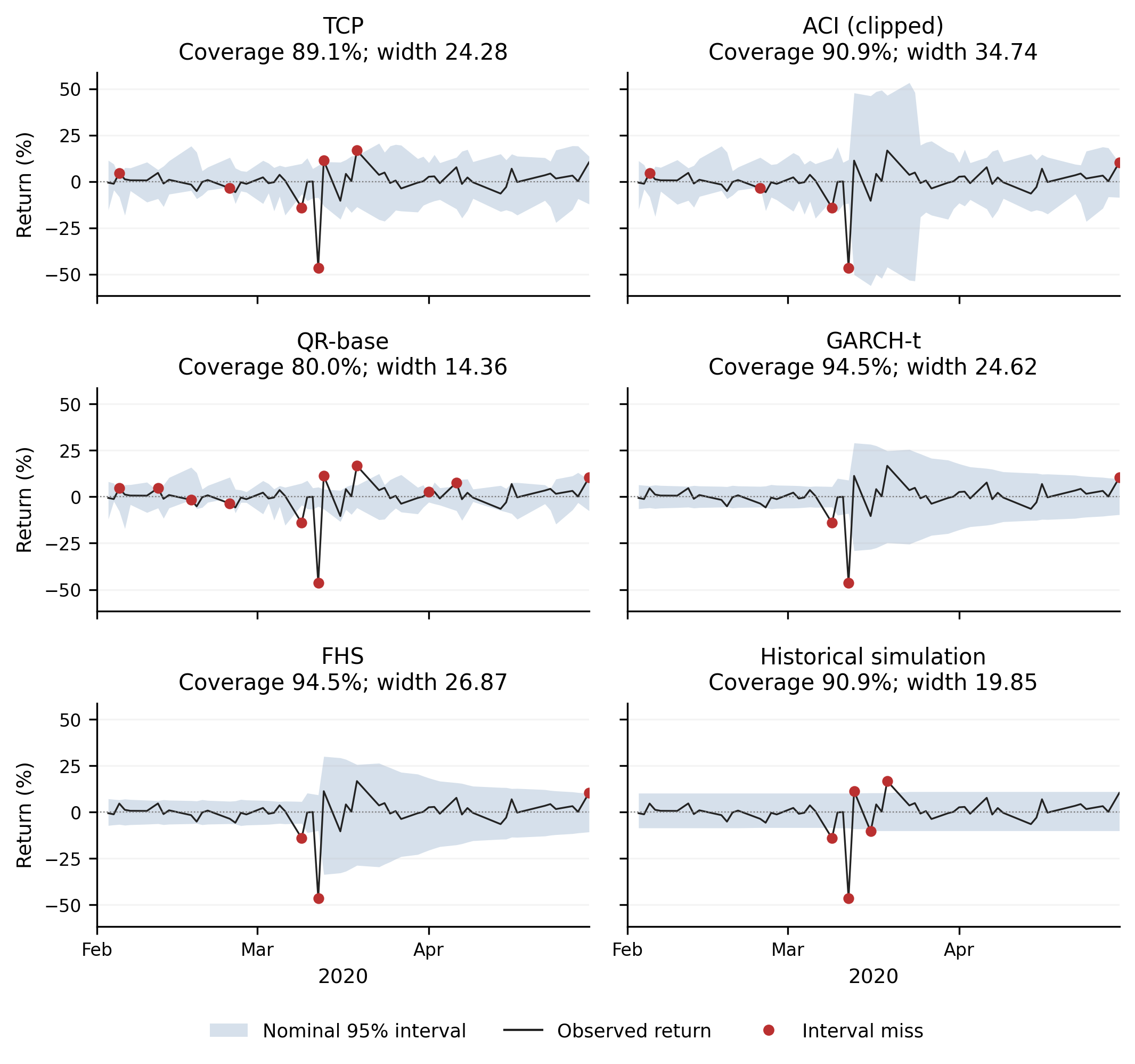}
\caption{Bitcoin prediction intervals from February to April 2020, with a nominal 95\% coverage target. Blue bands show intervals, black lines show observed returns, and red points mark interval misses. The six panels share their dates and vertical scale. Titles report coverage and mean width on the 55 recorded observations. Return values and widths use the percentage-point scale.}
\label{fig:crisis-bitcoin}
\end{figure}

\begin{figure}[p]
\centering
\includegraphics[width=\linewidth]{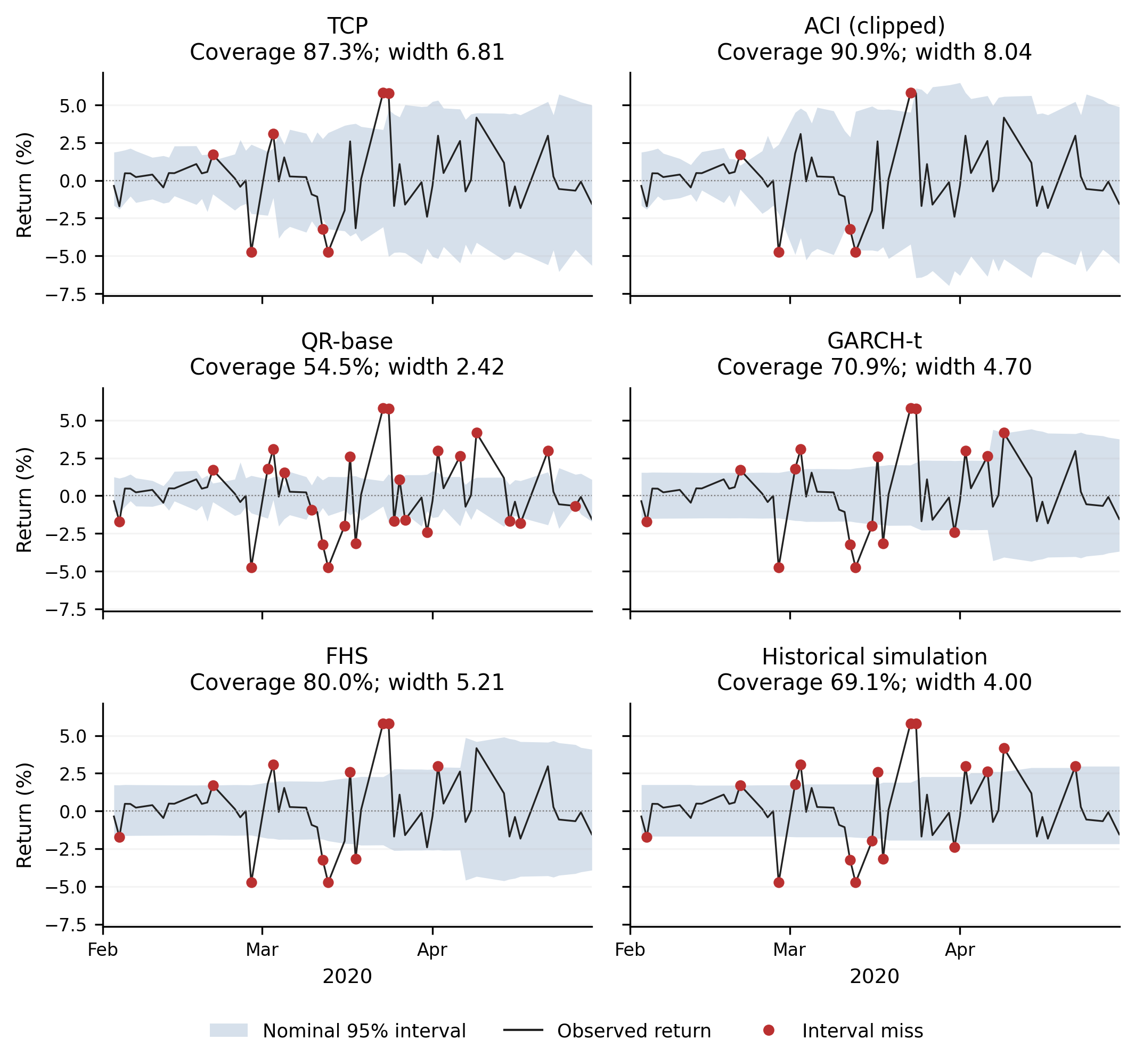}
\caption{Gold prediction intervals from February to April 2020. The display follows Figure~\ref{fig:crisis-bitcoin}. All six methods use the same 55 recorded observations and a nominal 95\% coverage target.}
\label{fig:crisis-gold}
\end{figure}

Tables~\ref{tab:covwidth-btc-crisis} and \ref{tab:covwidth-gold-crisis} report Bitcoin and Gold performance from February to April 2020. Each table uses 55 recorded observations. Figures~\ref{fig:crisis-bitcoin} and \ref{fig:crisis-gold} show the same six methods as Figure~\ref{fig:viz}. Figure~\ref{fig:tcp-tcprm-sp500} already gives the TCP and TCP-RM overlays for all three main assets.

On Bitcoin, TCP and TCP-RM both cover 0.8909 of the observations. GARCH-t and FHS each cover 0.9455, and RiskMetrics covers 0.9636. On Gold, TCP and TCP-RM both cover 0.8727. Clipped ACI covers 0.9091, while GARCH-t covers 0.7091. These are descriptive results from a short window, not evidence of uniform superiority across assets or market conditions.

\begin{table}[htbp]
\centering
\small
\caption{Bitcoin interval performance from February to April 2020. All methods use the same 55 recorded observations and a nominal 95\% coverage target. Width and the Winkler interval score are in percentage points.}
\label{tab:covwidth-btc-crisis}
\begin{tabular}{lrrr}
\toprule
Method & Coverage & Mean width & Winkler score \\
\midrule
TCP & 0.8909 & 24.2814 & 61.4155 \\
TCP-RM & 0.8909 & 24.2729 & 61.4293 \\
CQR & 0.8909 & 24.2814 & 61.4155 \\
ACI (clipped) & 0.9091 & 34.7380 & 65.3270 \\
QR-base & 0.8000 & 14.3566 & 72.5000 \\
GARCH-t & 0.9455 & 24.6180 & 58.6166 \\
FHS & 0.9455 & 26.8694 & 59.3985 \\
GARCH-EVT & 0.9455 & 28.1509 & 59.6640 \\
RiskMetrics & 0.9636 & 24.9727 & 59.2105 \\
Historical simulation & 0.9091 & 19.8499 & 57.9626 \\
\bottomrule
\end{tabular}
\end{table}

\begin{table}[htbp]
\centering
\small
\caption{Gold interval performance from February to April 2020. All methods use the same 55 recorded observations and a nominal 95\% coverage target. Width and the Winkler interval score are in percentage points.}
\label{tab:covwidth-gold-crisis}
\begin{tabular}{lrrr}
\toprule
Method & Coverage & Mean width & Winkler score \\
\midrule
TCP & 0.8727 & 6.8087 & 13.8130 \\
TCP-RM & 0.8727 & 6.8119 & 13.8119 \\
CQR & 0.8727 & 6.8087 & 13.8130 \\
ACI (clipped) & 0.9091 & 8.0437 & 11.8536 \\
QR-base & 0.5455 & 2.4158 & 30.0981 \\
GARCH-t & 0.7091 & 4.7006 & 19.2662 \\
FHS & 0.8000 & 5.2127 & 17.6184 \\
GARCH-EVT & 0.8000 & 5.1780 & 17.7713 \\
RiskMetrics & 0.8545 & 6.5266 & 13.3613 \\
Historical simulation & 0.6909 & 3.9957 & 20.0087 \\
\bottomrule
\end{tabular}
\end{table}

\clearpage
\section{Additional lower-tail backtests}
\label{app:backtests}

Tables~\ref{tab:backtests-bitcoin} and \ref{tab:backtests-gold} report the Bitcoin and Gold backtests. An exception is a return below the stated lower endpoint. It is not a miss on either side of a two-sided interval. For TCP, TCP-RM, CQR and the ACI constructions, the 2.5\% reference is an endpoint diagnostic. CQR-1s is constructed separately as a one-sided forecast.

Table~\ref{tab:tail-test-counts} counts the series with $p$-values strictly greater than 0.05. Its final column requires all three selected tests to satisfy that rule on the same series. Passing the unconditional-coverage test alone is not sufficient. The Gold results illustrate the distinction: FHS and GARCH-EVT pass unconditional and conditional coverage, but the dynamic-quantile test rejects both.

\begin{table}[htbp]
\centering
\small
\setlength{\tabcolsep}{4pt}
\caption{Bitcoin lower-tail backtests on 1,448 observations. Here $p$ is the reference exception probability. A dagger marks a two-sided interval endpoint used as a 2.5\% diagnostic. The two CQR-1s rows use the same forecasts, constructed at requested level $\tau=0.025$, with different reference probabilities. Test abbreviations follow Table~\ref{tab:backtests-sp500}.}
\label{tab:backtests-bitcoin}
\begin{tabular}{lrrrrrr}
\toprule
Forecast & $p$ & Exceptions & $p_{\rm UC}$ & $p_{\rm IND}$ & $p_{\rm CC}$ & $p_{\rm DQ}$ \\
\midrule
TCP$^\dagger$ & 0.025 & 27 & 0.1051 & 0.5283 & 0.2205 & 0.0063 \\
TCP-RM$^\dagger$ & 0.025 & 27 & 0.1051 & 0.5283 & 0.2205 & 0.0063 \\
CQR$^\dagger$ & 0.025 & 27 & 0.1051 & 0.5283 & 0.2205 & 0.0062 \\
ACI (clipped)$^\dagger$ & 0.025 & 31 & 0.3697 & 0.6944 & 0.6191 & 0.0240 \\
QR-base & 0.025 & 97 & $<10^{-4}$ & 0.8361 & $<10^{-4}$ & $<10^{-4}$ \\
GARCH-t & 0.025 & 38 & 0.7637 & 0.3616 & 0.6305 & 0.5747 \\
FHS & 0.025 & 35 & 0.8391 & 0.0590 & 0.1648 & 0.0588 \\
GARCH-EVT & 0.025 & 34 & 0.7084 & 0.2440 & 0.4730 & 0.3665 \\
RiskMetrics & 0.025 & 43 & 0.2661 & 0.1770 & 0.2166 & 0.2148 \\
Historical simulation & 0.025 & 38 & 0.7637 & 0.0929 & 0.2329 & 0.0043 \\
Reversed-sign control$^\dagger$ & 0.025 & 79 & $<10^{-4}$ & 0.4143 & $<10^{-4}$ & $<10^{-4}$ \\
Fixed-coefficient control & 0.025 & 90 & $<10^{-4}$ & 0.1542 & $<10^{-4}$ & $<10^{-4}$ \\
\midrule
CQR-1s & 0.025 & 27 & 0.1051 & 0.5283 & 0.2205 & 0.0018 \\
CQR-1s & $1/61$ & 27 & 0.5088 & 0.5283 & 0.6590 & $<10^{-4}$ \\
\bottomrule
\end{tabular}
\end{table}

\begin{table}[htbp]
\centering
\small
\setlength{\tabcolsep}{4pt}
\caption{Gold lower-tail backtests on 1,448 observations. Here $p$ is the reference exception probability. A dagger marks a two-sided interval endpoint used as a 2.5\% diagnostic. The two CQR-1s rows use the same forecasts, constructed at requested level $\tau=0.025$, with different reference probabilities. Test abbreviations follow Table~\ref{tab:backtests-sp500}.}
\label{tab:backtests-gold}
\begin{tabular}{lrrrrrr}
\toprule
Forecast & $p$ & Exceptions & $p_{\rm UC}$ & $p_{\rm IND}$ & $p_{\rm CC}$ & $p_{\rm DQ}$ \\
\midrule
TCP$^\dagger$ & 0.025 & 46 & 0.1131 & 0.2461 & 0.1455 & 0.0030 \\
TCP-RM$^\dagger$ & 0.025 & 46 & 0.1131 & 0.2461 & 0.1455 & 0.0030 \\
CQR$^\dagger$ & 0.025 & 46 & 0.1131 & 0.2461 & 0.1455 & 0.0031 \\
ACI (clipped)$^\dagger$ & 0.025 & 51 & 0.0188 & 0.1382 & 0.0211 & $<10^{-4}$ \\
QR-base & 0.025 & 136 & $<10^{-4}$ & 0.5747 & $<10^{-4}$ & $<10^{-4}$ \\
GARCH-t & 0.025 & 58 & 0.0007 & 0.6578 & 0.0030 & $<10^{-4}$ \\
FHS & 0.025 & 48 & 0.0583 & 0.7473 & 0.1582 & 0.0004 \\
GARCH-EVT & 0.025 & 44 & 0.2038 & 0.7530 & 0.4245 & 0.0007 \\
RiskMetrics & 0.025 & 57 & 0.0012 & 0.3348 & 0.0034 & 0.0004 \\
Historical simulation & 0.025 & 48 & 0.0583 & 0.7473 & 0.1582 & 0.0012 \\
Reversed-sign control$^\dagger$ & 0.025 & 125 & $<10^{-4}$ & 0.7882 & $<10^{-4}$ & $<10^{-4}$ \\
Fixed-coefficient control & 0.025 & 110 & $<10^{-4}$ & 0.5502 & $<10^{-4}$ & $<10^{-4}$ \\
\midrule
CQR-1s & 0.025 & 28 & 0.1508 & 0.5687 & 0.3029 & 0.0419 \\
CQR-1s & $1/61$ & 28 & 0.3909 & 0.5687 & 0.5884 & 0.0069 \\
\bottomrule
\end{tabular}
\end{table}

\begin{table}[htbp]
\centering
\small
\caption{Numbers of series with test $p$-values strictly greater than 0.05, out of thirteen series. The final column requires UC, CC and DQ all to exceed 0.05. A dagger marks a two-sided endpoint diagnostic. The two CQR-1s rows evaluate the same forecasts at different reference probabilities. These are descriptive test counts, not a joint test or proof of calibration.}
\label{tab:tail-test-counts}
\begin{tabular}{lrrrrr}
\toprule
Forecast & Reference $p$ & UC & CC & DQ & All three \\
\midrule
TCP$^\dagger$ & 0.025 & 11 & 11 & 1 & 1 \\
TCP-RM$^\dagger$ & 0.025 & 12 & 11 & 1 & 1 \\
CQR$^\dagger$ & 0.025 & 11 & 11 & 1 & 1 \\
ACI (clipped)$^\dagger$ & 0.025 & 10 & 9 & 1 & 1 \\
QR-base & 0.025 & 0 & 0 & 0 & 0 \\
GARCH-t & 0.025 & 8 & 8 & 10 & 7 \\
FHS & 0.025 & 13 & 13 & 12 & 12 \\
GARCH-EVT & 0.025 & 13 & 13 & 11 & 11 \\
RiskMetrics & 0.025 & 4 & 7 & 5 & 3 \\
Historical simulation & 0.025 & 9 & 7 & 3 & 3 \\
Reversed-sign control$^\dagger$ & 0.025 & 0 & 0 & 0 & 0 \\
Fixed-coefficient control & 0.025 & 0 & 0 & 0 & 0 \\
\midrule
CQR-1s & 0.025 & 8 & 9 & 4 & 3 \\
CQR-1s & $1/61$ & 13 & 13 & 2 & 2 \\
\bottomrule
\end{tabular}
\end{table}

\clearpage
\section{Additional sensitivity results}
\label{app:sensitivity}

Tables~\ref{tab:sensitivity-bitcoin} and \ref{tab:sensitivity-gold} give the Bitcoin and Gold grid. They include TCP and TCP-RM at both calibration sizes, $m=40$ and $m=60$. The decay parameters remain $\lambda=0.1$ and $\beta=0.7$. The main configuration is $w=252$, $m=60$ and $\gamma_0=0.01$.

Across the TCP-RM grid, Bitcoin coverage ranges from 0.9537 to 0.9608. Gold coverage ranges from 0.9425 to 0.9542. These ranges describe the configurations evaluated. They do not establish robustness beyond this grid or show that the offset always improves performance.

The training size is $w-m$. The three window sizes also give different sample lengths: 1,600, 1,448 and 1,200 observations for $w=100$, 252 and 500. Window comparisons therefore combine changes in fitting history with changes in evaluation dates. Step-size comparisons within a fixed $(w,m)$ pair use the same dates and base forecasts.

\begin{table}[htbp]
\centering
\small
\caption{Bitcoin sensitivity to total window $w$, calibration size $m$ and initial step size $\gamma_0$. A dash denotes TCP without the online offset; the other rows use TCP-RM. The training size is $w-m$. Width and the Winkler score are in percentage points. Sample size $n$ varies with $w$.}
\label{tab:sensitivity-bitcoin}
\begin{tabular}{rrrrrrr}
\toprule
$w$ & $m$ & $\gamma_0$ & $n$ & Coverage & Mean width & Winkler score \\
\midrule
100 & 40 & -- & 1600 & 0.9575 & 21.7792 & 29.3167 \\
100 & 40 & 0.005 & 1600 & 0.9575 & 21.7806 & 29.3170 \\
100 & 40 & 0.010 & 1600 & 0.9575 & 21.7819 & 29.3174 \\
100 & 40 & 0.050 & 1600 & 0.9575 & 21.7926 & 29.3202 \\
\midrule
100 & 60 & -- & 1600 & 0.9544 & 21.3281 & 28.5659 \\
100 & 60 & 0.005 & 1600 & 0.9544 & 21.3311 & 28.5664 \\
100 & 60 & 0.010 & 1600 & 0.9544 & 21.3342 & 28.5668 \\
100 & 60 & 0.050 & 1600 & 0.9544 & 21.3585 & 28.5703 \\
\midrule
252 & 40 & -- & 1448 & 0.9572 & 21.6121 & 28.9690 \\
252 & 40 & 0.005 & 1448 & 0.9572 & 21.6138 & 28.9693 \\
252 & 40 & 0.010 & 1448 & 0.9572 & 21.6154 & 28.9697 \\
252 & 40 & 0.050 & 1448 & 0.9572 & 21.6285 & 28.9722 \\
\midrule
252 & 60 & -- & 1448 & 0.9537 & 20.8879 & 27.9846 \\
252 & 60 & 0.005 & 1448 & 0.9537 & 20.8869 & 27.9849 \\
252 & 60 & 0.010 & 1448 & 0.9537 & 20.8859 & 27.9851 \\
252 & 60 & 0.050 & 1448 & 0.9537 & 20.8784 & 27.9871 \\
\midrule
500 & 40 & -- & 1200 & 0.9608 & 20.5048 & 27.1633 \\
500 & 40 & 0.005 & 1200 & 0.9608 & 20.5050 & 27.1637 \\
500 & 40 & 0.010 & 1200 & 0.9608 & 20.5053 & 27.1640 \\
500 & 40 & 0.050 & 1200 & 0.9608 & 20.5071 & 27.1666 \\
\midrule
500 & 60 & -- & 1200 & 0.9550 & 20.1261 & 27.0874 \\
500 & 60 & 0.005 & 1200 & 0.9550 & 20.1365 & 27.0890 \\
500 & 60 & 0.010 & 1200 & 0.9550 & 20.1470 & 27.0907 \\
500 & 60 & 0.050 & 1200 & 0.9550 & 20.2307 & 27.1040 \\
\bottomrule
\end{tabular}
\end{table}

\begin{table}[htbp]
\centering
\small
\caption{Gold sensitivity to total window $w$, calibration size $m$ and initial step size $\gamma_0$. A dash denotes TCP without the online offset; the other rows use TCP-RM. The training size is $w-m$. Width and the Winkler score are in percentage points. Sample size $n$ varies with $w$.}
\label{tab:sensitivity-gold}
\begin{tabular}{rrrrrrr}
\toprule
$w$ & $m$ & $\gamma_0$ & $n$ & Coverage & Mean width & Winkler score \\
\midrule
100 & 40 & -- & 1600 & 0.9487 & 4.6414 & 6.0105 \\
100 & 40 & 0.005 & 1600 & 0.9494 & 4.6437 & 6.0107 \\
100 & 40 & 0.010 & 1600 & 0.9494 & 4.6460 & 6.0110 \\
100 & 40 & 0.050 & 1600 & 0.9494 & 4.6643 & 6.0133 \\
\midrule
100 & 60 & -- & 1600 & 0.9487 & 4.7256 & 6.2217 \\
100 & 60 & 0.005 & 1600 & 0.9487 & 4.7304 & 6.2219 \\
100 & 60 & 0.010 & 1600 & 0.9487 & 4.7351 & 6.2221 \\
100 & 60 & 0.050 & 1600 & 0.9500 & 4.7687 & 6.2241 \\
\midrule
252 & 40 & -- & 1448 & 0.9454 & 4.7756 & 6.1181 \\
252 & 40 & 0.005 & 1448 & 0.9461 & 4.7812 & 6.1180 \\
252 & 40 & 0.010 & 1448 & 0.9461 & 4.7868 & 6.1180 \\
252 & 40 & 0.050 & 1448 & 0.9468 & 4.8285 & 6.1185 \\
\midrule
252 & 60 & -- & 1448 & 0.9427 & 4.7368 & 6.3246 \\
252 & 60 & 0.005 & 1448 & 0.9427 & 4.7403 & 6.3246 \\
252 & 60 & 0.010 & 1448 & 0.9427 & 4.7438 & 6.3246 \\
252 & 60 & 0.050 & 1448 & 0.9434 & 4.7673 & 6.3251 \\
\midrule
500 & 40 & -- & 1200 & 0.9492 & 5.0928 & 6.5739 \\
500 & 40 & 0.005 & 1200 & 0.9500 & 5.1058 & 6.5748 \\
500 & 40 & 0.010 & 1200 & 0.9500 & 5.1188 & 6.5758 \\
500 & 40 & 0.050 & 1200 & 0.9542 & 5.2031 & 6.5862 \\
\midrule
500 & 60 & -- & 1200 & 0.9425 & 4.9656 & 6.6141 \\
500 & 60 & 0.005 & 1200 & 0.9425 & 4.9933 & 6.6114 \\
500 & 60 & 0.010 & 1200 & 0.9442 & 5.0170 & 6.6098 \\
500 & 60 & 0.050 & 1200 & 0.9517 & 5.1917 & 6.6117 \\
\bottomrule
\end{tabular}
\end{table}

\clearpage
\section{TCP and TCP-RM traces}
\label{app:traces}

\begin{figure}[p]
\centering
\includegraphics[width=\linewidth,height=0.78\textheight,keepaspectratio]{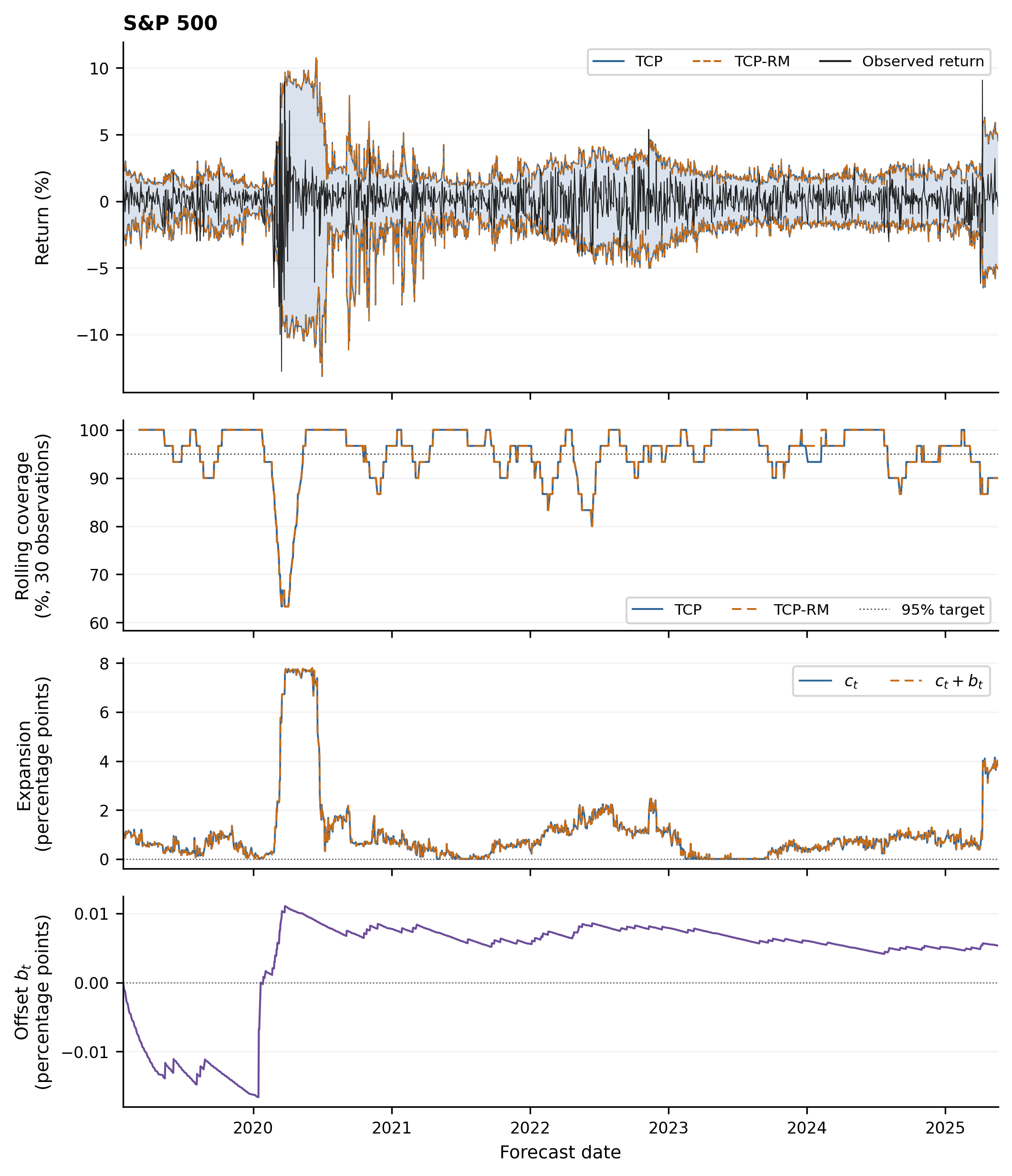}
\caption{S\&P 500 TCP and TCP-RM traces on the 1,448 forecast dates. From top to bottom: observed returns and nominal 95\% intervals; coverage over the latest 30 recorded observations, with the 95\% target marked; the TCP expansion $c_t$ and TCP-RM expansion $c_t+b_t$; and the offset $b_t$ on a separate scale. The shaded band is the TCP interval.}
\label{fig:trace-sp500}
\end{figure}

\begin{figure}[p]
\centering
\includegraphics[width=\linewidth,height=0.78\textheight,keepaspectratio]{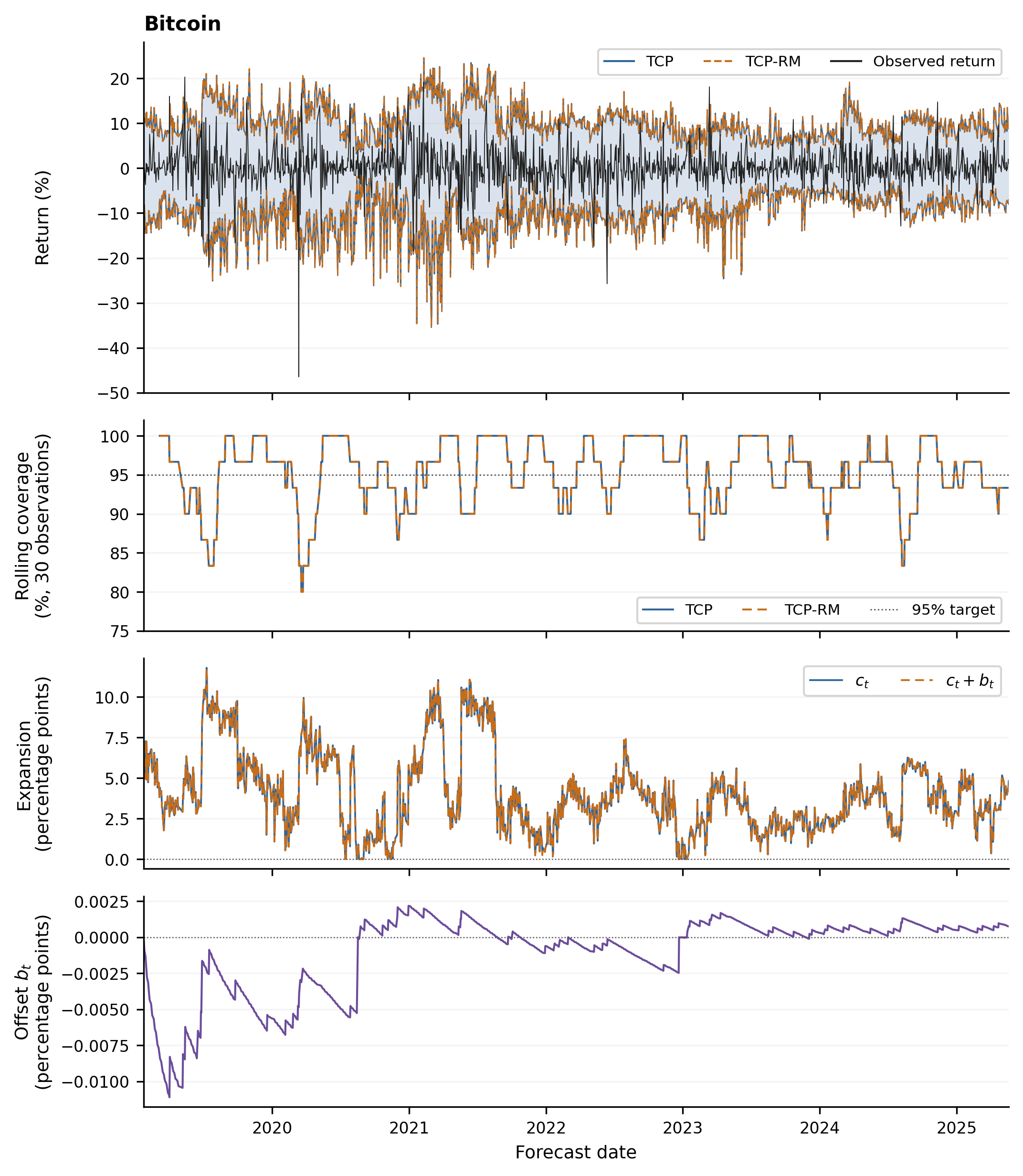}
\caption{Bitcoin TCP and TCP-RM traces. The four panels follow Figure~\ref{fig:trace-sp500}. The rolling window counts recorded observations; it is not a 30-calendar-day window.}
\label{fig:trace-bitcoin}
\end{figure}

\begin{figure}[p]
\centering
\includegraphics[width=\linewidth,height=0.78\textheight,keepaspectratio]{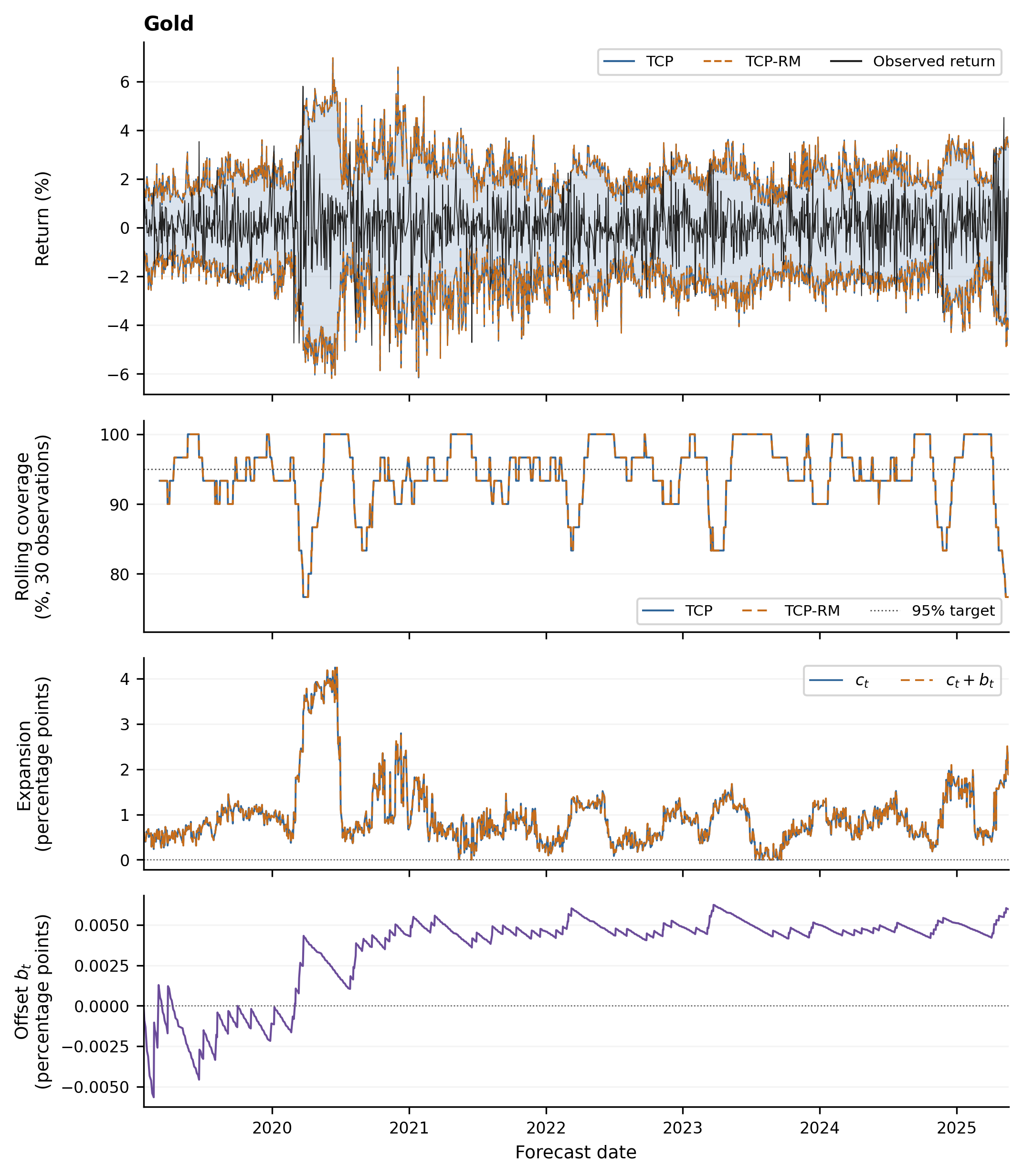}
\caption{Gold TCP and TCP-RM traces. The four panels follow Figure~\ref{fig:trace-sp500}. The offset is shown separately from the total expansion so that its scale and sign remain visible.}
\label{fig:trace-gold}
\end{figure}

Figures~\ref{fig:trace-sp500}--\ref{fig:trace-gold} show the full evaluation period. The rolling coverage at date $t$ is
\[
 \widehat{\mathrm{Cov}}_{30,t}
 =\frac1{30}\sum_{i=t-29}^{t}
       \mathbf{1}\{\ell_i\le r_i\le u_i\}.
\]
The window counts recorded observations. Values are shown only after 30 forecasts are available. Each rolling value includes the outcome at its plotted date, so it is a retrospective diagnostic rather than an issued forecast or a conditional-coverage guarantee.

The expansion and offset are recovered from the forecast endpoints:
\[
 c_t=u_t^{\mathrm{TCP}}-u_t^{\mathrm{QR}},\qquad
 b_t=u_t^{\mathrm{RM}}-u_t^{\mathrm{TCP}}.
\]
The corresponding lower-endpoint differences give the same quantities to numerical precision. The recovered offsets also satisfy \eqref{eq:rm-update} to numerical precision. We plot $b_t$ on its own axis because it is small relative to $c_t$ in this experiment. The interval and expansion panels can therefore show nearly overlapping curves even when the offset changes.

\end{document}